\documentclass[10pt,a4paper,twoside,english]{elsarticle}
\usepackage[latin1]{inputenc}
\usepackage{babel}
\usepackage{array}
\usepackage{verbatim}
\usepackage{rotfloat}
\usepackage{url}
\usepackage{multirow}
\usepackage{amsmath}
\usepackage{amssymb}
\usepackage{graphicx}
\usepackage{nomencl}
\providecommand{\printnomenclature}{\printglossary}
\providecommand{\makenomenclature}{\makeglossary}
\makenomenclature
\usepackage[unicode=true,
 bookmarks=false,
 breaklinks=false,pdfborder={0 0 1},backref=section,colorlinks=false]
 {hyperref}

\makeatletter

\DeclareTextSymbolDefault{\textquotedbl}{T1}
\providecommand{\tabularnewline}{\\}

\@ifundefined{date}{}{\date{}}
\def\ps@pprintTitle{%
 \let\@oddhead\@empty
 \let\@evenhead\@empty
 \def\@oddfoot{\centerline{\thepage}}%
 \let\@evenfoot\@oddfoot}

\makeatother

\begin{document}

\begin{frontmatter}{}

\title{Eigen Artificial Neural Networks}

\author[FYB]{Francisco Yepes Barrera}

\cortext[FYB]{paco.yepes@godelia.org}
\begin{abstract}
This work has its origin in intuitive physical and statistical considerations.
The problem of optimizing an artificial neural network is treated
as a physical system, composed of a conservative vector force field.
The derived scalar potential is a measure of the potential energy
of the network, a function of the distance between predictions and
targets.

Starting from some analogies with wave mechanics, the description
of the system is justified with an eigenvalue equation that is a variant
of the Schrõdinger equation, in which the potential is defined by
the mutual information between inputs and targets. The weights and
parameters of the network, as well as those of the state function,
are varied so as to minimize energy, using an equivalent of the variational
theorem of wave mechanics. The minimum energy thus obtained implies
the principle of minimum mutual information (MinMI). We also propose
a definition of the potential work produced by the force field to
bring a network from an arbitrary probability distribution to the
potential-constrained system, which allows to establish a measure
of the complexity of the system. At the end of the discussion we expose
a recursive procedure that allows to refine the state function and
bypass some initial assumptions, as well as a discussion of some topics
in quantum mechanics applied to the formalism, such as the uncertainty
principle and the temporal evolution of the system.

Results demonstrate how the minimization of energy effectively leads
to a decrease in the average error between network predictions and
targets.
\end{abstract}

\end{frontmatter}{}

\section*{A note on the symbolism}

In the continuation of this work we will represent in bold the set
of independent components of a variable and in italic with subscripts
the operations on the single components. In the paper considerations
are made and a model is constructed starting from the expected values
(mean) of quantities, modeled with appropriate probability densities,
and not on the concrete measurements of these quantities. Considering
a generic multivariate variable $Q$, $\mathbf{q}$ represents the
set of components of $Q$ (number of features) while $q_{i}$ represents
each component being a vector composed of the individual measurements
$q_{ij}$. Similarly, $d\mathbf{x}=\prod_{i}dx_{i}$  and $d\mathbf{t}=\prod_{k}t_{k}$
represent respectively the differential element of volume in the space
of inputs and targets, whereby $\int(\ldots)\,d\mathbf{x}$ and $\int(\ldots)\,d\mathbf{t}$
express the integration on the whole space while $\int(\ldots)\,dx_{i}$
and $\int(\ldots)\,dt_{k}$ the integration on the individual components.
In the text, the statistical modeling takes place at the level of
the individual components, which in the case of inputs and targets
is constructed with probability densities generated from the relative
measurements. $\left\langle \ldots\right\rangle $ means expected
value.

When not specified, we will implicitly assume that integrals extend
to the whole space in the interval $[-\infty:\infty]$.

\section{\label{sec:Introduzione}Introduction}

This paper analyzes the problem of optimizing artificial neural networks
(ANNs), ie the problem of finding functions $y(\mathbf{x};\Gamma)$,
dependent on matrices of input data $\mathbf{x}$ and parameters $\Gamma$,
such that, given a target $\mathbf{t}$ make an optimal mapping between
$\mathbf{x}$ and $\mathbf{t}$ \citep{husmeier_modelling_1997}.
The treatment is limited to pattern recognition problems related to
functional approximation of continuous variables, excluding time series
analysis and pattern recognition problems involving discrete variables.

The starting point of this paper is made up of some well-known theoretical
elements:
\begin{enumerate}
\item Generally, the training of an artificial neural network consists in
the minimization of some error function between the output of the
network $y(\mathbf{x};\Gamma)$ and the target $\mathbf{t}$. In the
best case it identify the global minimum of the error; in general
it finds local minima. The total of minima compose a discrete set
of values.
\item The passage from a prior to a posterior or conditional probability,
that is the observation or acquisition of additional knowledge about
data, implies a collapse of the function that describes the system:
the conditional probability calculated with Bayes' theorem leads to
distributions of closer and more localized probabilities than prior
ones \citep{Bishop:NNPR:1995}.
\item Starting from the formulation of the mean square error produced by
an artificial neural network and considering a set $C$ of targets
$t_{k}$ whose distributions are independent

\begin{equation}
p(\mathbf{t}|\mathbf{x})=\prod_{k=1}^{C}p(t_{k}|\mathbf{x})\label{eq:p(t|x) indipendenti}
\end{equation}
\begin{equation}
p(\mathbf{t})=\prod_{k=1}^{C}p(t_{k})\label{eq:p(t) indipendenti}
\end{equation}
where $p(t_{k}|\mathbf{x})$ is the conditional probability of $t_{k}$
given $\mathbf{x}$ and $p(t_{k})$ is the marginal probability of
$t_{k}$, it can be shown that
\begin{equation}
\left\langle (y_{k}-t_{k})^{2}\right\rangle \geq\left\langle (\left\langle t_{k}|\mathbf{x}\right\rangle -t_{k})^{2}\right\rangle \label{eq:variational theorem ANNs}
\end{equation}
being $\left\langle t_{k}|\mathbf{x}\right\rangle $ the expected
value or conditional average of $t_{k}$ given $\mathbf{x}$, and
the equal valid only at the optimum. In practice, any trial function
$y_{k}(\mathbf{x};\Gamma)$ leads to a quadratic deviation respect
to $t_{k}$ greater than that generated by the optimal function, $y'_{k}=y_{k}(\mathbf{x};\Gamma')$,
corresponding to the absolute minimum of the error, since it represents
the conditional average of the target, as demonstrated by the known
result \citep{Bishop:NNPR:1995}
\end{enumerate}
\begin{equation}
y'_{k}=\left\langle t_{k}|\mathbf{x}\right\rangle \label{eq:ouput rete neurale come media condizionale}
\end{equation}

These points have analogies with three theoretical elements underlying
wave mechanics \citep{levine_quantum_2014}:
\begin{enumerate}
\item Any physical system described by the Schrõdinger equation and constrained
by a scalar potential $V(\mathbf{x})$ leads to a quantization of
the energy values, which constitute a discrete set of real values.
\item A quantum-mechanical system is composed by the superposition of several
states described by the Schrõdinger equation, corresponding to as
many eigenvalues. The observation of the system causes the collapse
of the wave function on one of the states, being only possible to
calculate the probability of obtaining the different eigenvalues.
\item \label{enu:analogies with eave mechanics, point 3}When it is not
possible to analytically obtain the true wave function $\Psi'$ and
the true energy $E'$ of a quantum-mechanical system, it is possible
to use trial functions $\Psi$, with eigenvalues $E$, dependent on
a set $\Gamma$ of parameters. In this case we can find an approximation
to $\Psi'$ and $E'$ varying $\Gamma$ and taking into account the
variational theorem
\[
\left\{ E=\frac{\int\Psi{}^{*}\hat{H}\Psi\,d\mathbf{x}}{\int\Psi{}^{*}\Psi\,d\mathbf{x}}\right\} \geq E'
\]
\end{enumerate}
Regarding point \ref{enu:analogies with eave mechanics, point 3},
we can consider the condition (\ref{eq:variational theorem ANNs})
as an equivalent of the variational theorem for artificial neural
networks.

\section{\label{sec:Quantum-mechanics-and information theory}Quantum mechanics
and information theory}

The starting point of this work consists of some analogies and intuitive
considerations. Similar analogies based on the structural similarity
between the equations that govern some phenomena and Schrödinger's
equation are found in other research areas. For example, the Black-Scholes
equation for the pricing of options can be interpreted as the Schrödinger
equation of the free particle \citep{contreras_quantum_2010,romero_black-scholes_2011}.

As we will see in the part related to the analysis of the results,
empirical evidence in applying the model to some datasets supports
the validity of the mathematical formalism detailed in the following
sections. However, the premises of this paper suggest a relationship
between quantum mechanics and information theory without adequate
justification. The rich literature available on this topic mitigates
this inadequacy. We will not do a thorough analysis on the matter,
but it is worth mentioning some significant works:
\begin{enumerate}
\item Kurihara and Uyen Quach \citep{kurihara_advantages_2015} analyzed
the relationship between probability amplitude in quantum mechanics
and information theory.
\item Stam \citep{stam_inequalities_1959} and Hradil and \v Reh\'a\v cek
\citep{hradil_uncertainty_2004} highlighted the link between the
uncertainty relationships in quantum mechanics, Shannon's entropy
and Fisher's information. This last quantity is related to the Cramér-Rao
bound, from which it is possible to derive the uncertainty principle
\citep{angelow_evolution_2009,dembo_information_1991,frowis_tighter_2015,gibilisco_robertson-type_2008,gibilisco_uncertainty_2007,hall_prior_2004,parthasarathy_philosophy_2009,rodriguez_disturbance-disturbance_2018,theodoridis_machine_2015}.
The relationship between Schrödinger's equation and Fisher's information
has also been studied by Plastino \citep{deloumeaux_information_2012},
Fischer \citep{fischer_limiting_2019} and Flego et al. \citep{flego_fisher_2012}.
Frank and Lieb \citep{frank_entropy_2012} have shown how the diagonal
elements of the density matrix determine classical entropies whose
sum is a measure of the uncertainty principle. Falaye et al. \citep{falaye_fishers_2016}
analyzed the quantum-mechanical probability for ground and excited
states through Fisher information, together with the relationship
between the uncertainty principle and Cramér-Rao bound. Entropic uncertainty
relations have been discussed by Bialynicki-Birula and Mycielski \citep{bialynicki-birula_uncertainty_1975},
Coles et al. \citep{coles_entropic_2017} and Hilgevoord and Uffink
\citep{hilgevoord_uncertainty_2016}.
\item Reginatto showed the relationship between information theory and the
hydrodynamic formulation of quantum mechanics \citep{reginatto_hydrodynamical_1999}.
\item Braunstein \citep{braunstein_wringing_1990} and Cerf and Adami \citep{cerf_entropic_1997}
showed how Bell's inequality can be written in terms of the average
information and analyzed from an information-theoretic point of view.
\item Parwani \citep{parwani_why_2005} highlights how a link between the
linearity of the Schrödinger equation and the Lorentz invariance can
be obtained starting from information-theoretic arguments. Yuan and
Parwani \citep{yuan_properties_2009} have also demonstrated within
nonlinear Schrödinger equations how the criterion of minimizing energy
is equivalent to maximizing uncertainty, understood in the sense of
information theory.
\item Klein \citep{pahlavani_statistical_2012,klein_probabilistic_2019}
showed how Schrödinger's equation can be derived from purely statistical
considerations, showing that quantum theory is a substructure of classical
probabilistic physics.
\end{enumerate}

\section{\label{sec:Treatment-as eingevalues of ANNs}Treatment of the optimization
of artificial neural networks as an eigenvalue problem}

The analogies highlighted in Section \ref{sec:Introduzione} suggest
the possibility of dealing with the problem of optimizing artificial
neural networks as a physical system.\footnote{In the scientific literature it is possible to find interesting studies
of the application of mathematical physics equations to artificial
intelligence \citep{movellan_learning_1993}.} These analogies, of course, are not sufficient to justify the treatment
of the problem with eigenvalue equations, as happens in the physical
systems modeled by the Schrödinger equation, and are used in this
paper exclusively as a starting point that deserves further study.
However it is a line of research that can clarify intimate aspects
of the optimization of an artificial neural network and propose a
new point of view of this process. We will demonstrate in the following
sections that meaningful conclusions can be reached and that the proposed
treatment actually allows to optimize artificial neural networks by
applying the formalism to some datasets available in literature. A
first thought on the model is that it allows to naturally define the
energy of the network, a concept already used in some types of ANNs,
such as the Hopfield networks in which Lyapunov or energy functions
can be derived for binary element networks allowing a complete characterization
of their dynamics, and permits to generalize the concept of energy
for any type of ANN.

Suppose we can define a conservative force generated by the set of
targets $\mathbf{t}$, represented in the input space $\mathbf{x}$
with a vector field, being $N$ the dimension of $\mathbf{x}$. In
this case we have a scalar function $V(\mathbf{x})$, called potential,
which depends exclusively on the \emph{position}\footnote{In this discussion, ``position'' means the location of a point in
the input space $\mathbf{x}$.} an that is related to the force as

\begin{equation}
\mathbf{F}=-\nabla V(\mathbf{x})\label{eq:Force}
\end{equation}
which implies that the potential of the force at a point is proportional
to the potential energy possessed by an object at that point due to
the presence of the force. The negative sign in the equation (\ref{eq:Force})
means that the force is directed towards the target, where force and
potential are minimal, so $\mathbf{t}$ generates a force that \emph{attracts}
the \emph{bodies} immersed in the field, represented by the average
predictions of the network, with an intensity proportional to some
function of the distance between $y(\mathbf{x};\Gamma)$ and $\mathbf{t}$.
Greater is the error committed by a network in the target prediction,
greater is the potential energy of the system, which generates a increase
of the force that attracts the system to optimal points, represented
by networks whose parameterization allows to obtain predictions of
the target with low errors.

Equation (\ref{eq:ouput rete neurale come media condizionale}) highlights
how, at the optimum, the output of an artificial neural network is
an approximation to the conditional averages or expected values of
the targets $\mathbf{t}$ given the input $\mathbf{x}$. For a problem
of functional approximation like those analyzed in this paper, both
$\mathbf{x}$ and $\mathbf{t}$ are given by a set of measurements
for the problem (dataset), with average values that do not vary over
time. We therefore hypothesize a stationary system and an time independent
eigenvalue equation, having the same structure as the Schrõdinger
equation
\begin{equation}
-\epsilon\nabla^{2}\Psi(\mathbf{x})+V(\mathbf{x})\Psi(\mathbf{x})=E\Psi(\mathbf{x})\label{eq:State equation 1}
\end{equation}
where $\Psi$ is the state function of the system (network), $V$
a scalar potential, $E$ the network energy, and $\epsilon$ a multiplicative
constant. Given the mathematical structure of the model, we will refer
to the systems obtained from equation (\ref{eq:State equation 1})
as \emph{Eigen Artificial Neural Networks} (EANNs).

Equation (\ref{eq:State equation 1}) implements a parametric model
for the ANNs in which the optimization consists in minimizing, on
average, the energy of the network, function of $y(\mathbf{x};\Gamma)$
and $\mathbf{t}$, modeled by appropriate probability densities and
a set of variational parameters $\Gamma$. The working hypothesis
is that the minimization of energy through a parameter-dependent trial
function that makes use of the variational theorem (\ref{eq:variational theorem ANNs})
leads, using an appropriate potential, to a reduction of the error
committed by the network in the prediction of $\mathbf{t}$. In the
continuation of this paper we will consider the system governed by
the equation (\ref{eq:State equation 1}) a quantum system in all
respects, and we will implicitly assume the validity and applicability
of the laws of quantum mechanics.

\section{The potential\label{sec:The-potential}}

Globerson et al. \citep{Globerson:2009:MIPNCA} have studied the stimulus-response
relationship in neural populations activity trying to understand what
is the amount of information transmitted and have proposed the principle
of minimum mutual information (MinMI) between stimulus and response,
which we will consider valid in the context of artificial neural networks
and we will use in the variant of the relationship between $\mathbf{x}$
and $\mathbf{t}$.\footnote{In the next, we will use the proposal by Globerson et al. translating
it into the symbolism used in this paper.} An analog principle, from a point of view closer to information theory,
is given by Chen et al. \citep{chen_adaptive_2008,chen_system_2013},
who minimize the error/input information (EII) corresponding to the
mutual information (MI) between the identification error (a measure
of the difference between model and target, $\mathbf{t}-\mathbf{y}$)
and input $\mathbf{x}$. The intuitive idea behind the proposal of
Globerson et al. is that, among all the systems consistent with the
partial measured data of $\mathbf{x}$ and $\mathbf{t}$ (ie all the
systems $\mathbf{y}$ that differ in the set of parameters $\Gamma$),
the nearest one to the true relationship between stimulus and response
is given by the system with lower mutual information, since the systems
with relatively high MI contain an additional source of information
(noise) while the one with minimal MI contains the information that
can be attributed principally only to the measured data and further
simplification in terms of MI is not possible. An important implication
of this construction is that the MI of the true system (the system
in the limit of an infinite number of observations of $\mathbf{x}$
and $\mathbf{t}$) is greater than or equal to the minimum MI possible
between all systems consistent with the observations, since the true
system will contain an implicit amount of noise that can only be greater
than or equal to that of the system with minimum MI.\footnote{The work of Globerson et al. is proposed in the context of neural
coding. Here we give an interpretation so as to allow a coherent integration
within the issues related to the optimization of artificial neural
networks.}

So, a function that seems to be a good candidate to be used as potential
is the mutual information \citep{Xu:1999} between input and target,
$I(\mathbf{t},\mathbf{x})$, which is a positive quantity and whose
minimum corresponds to the minimum potential energy, and therefore
to the minimum Kullback-Leibler divergence between the joint probability
density of target $\mathbf{t}$ and input $\mathbf{x}$ and the marginal
probabilities $p(\mathbf{t})$ and $p(\mathbf{x})$
\begin{equation}
I(\mathbf{t},\mathbf{x})=\iint p(\mathbf{t}|\mathbf{x})p(\mathbf{x})\ln\left(\frac{p(\mathbf{t}|\mathbf{x})}{p(\mathbf{t})}\right)\,d\mathbf{x}d\mathbf{t}\label{eq:I(t,x)}
\end{equation}
In this case, the minimization of energy through a variational state
function that satisfies the equation (\ref{eq:State equation 1})
implies the principle of minimum mutual information \citep{fitzgerald_minimal_2011,globerson_minimum_2004,Globerson:2009:MIPNCA,zhang_information_2017},
equivalent to the principle of maximum entropy (\emph{MaxEnt}) \citep{finnegan_maximum_2017,park_maximum_1997,pires_minimum_2012,xiaodong_evaluation_2014}
when, as is our case, the marginal densities are known. In the following,
in order to make explicit the functional dependence in the expressions
of the integrals, we will call $I_{k}$ the function inside the integral
in equation (\ref{eq:I(t,x)}) relative to a single target $t_{k}$
\[
I_{k}(t_{k},\mathbf{x})=p(t_{k}|\mathbf{x})p(\mathbf{x})\ln\left(\frac{p(t_{k}|\mathbf{x})}{p(t_{k})}\right)
\]

The scalar potential depends only on the position $\mathbf{x}$ and
for $C$ targets becomes
\begin{equation}
V(\mathbf{x})=\sum_{k=1}^{C}\int I_{k}(t_{k},\mathbf{x})\,dt_{k}\label{eq:V(x) integral form}
\end{equation}
The equation (\ref{eq:V(x) integral form}) assumes a superposition
principle, similar to the valid one in the electric field, in which
the total potential is given by the sum of the potentials of each
of the $C$ targets of the problem, which implies that the field generated
by each target $t_{k}$ is independent of the field generated by the
other targets.\footnote{This superposition principle, in which the total field generated by
a set of sources is equal to the sum of the single fields produced
by each source, has no relation to the principle of superposition
of states in quantum mechanics.} In fact it can be shown that this principle is a direct consequence
of the independence of the densities (\ref{eq:p(t|x) indipendenti})
and (\ref{eq:p(t) indipendenti})
\begin{equation}
\begin{array}{l}
\int p(\mathbf{t}|\mathbf{x})p(\mathbf{x})\ln\left(\frac{p(\mathbf{\mathbf{t}}|\mathbf{x})}{p(\mathbf{t})}\right)\,d\mathbf{t}\\
=\int p(\mathbf{x})\ln\left(\prod_{k=1}^{C}\frac{p(t_{k}|\mathbf{x})}{p(t_{k})}\right)\prod_{j=1}^{C}p(t_{j}|\mathbf{x})\,d\mathbf{t}\\
=\sum_{k=1}^{C}\int p(\mathbf{x})\ln\left(\frac{p(t_{k}|\mathbf{x})}{p(t_{k})}\right)\prod_{j=1}^{C}p(t_{j}|\mathbf{x})\,d\mathbf{t}\\
=\sum_{k=1}^{C}\int p(t_{k}|\mathbf{x})p(\mathbf{x})\ln\left(\frac{p(t_{k}|\mathbf{x})}{p(t_{k})}\right)\,dt_{k}\int\prod_{j\neq k}^{C}p(t_{j}|\mathbf{x})\,d\mathbf{t}_{j\neq k}\\
=\sum_{k=1}^{C}\int p(t_{k}|\mathbf{x})p(\mathbf{x})\ln\left(\frac{p(t_{k}|\mathbf{x})}{p(t_{k})}\right)\,dt_{k}
\end{array}\label{eq:superposition principle}
\end{equation}
where $\int\prod_{j\neq k}^{C}p(t_{j}|\mathbf{x})\,d\mathbf{t}_{j\neq k}=\prod_{j\neq k}^{C}\int p(t_{j}|\mathbf{x})\,dt_{j}=1$
in the case of normalized probability densities.\footnote{All probability densities used in this paper are normalized.}
This result has a general character and is independent of the specific
functional form given to the probabilities $p(t_{k}|\mathbf{x})$
and $p(t_{k})$.

The conservative field proposed in this paper and the trend for the
derived potential and force suggests a qualitative analogy with the
physical mechanism of the harmonic oscillator, which in the one-dimensional
case has a potential $V(x)=\frac{1}{2}kx^{2}$, which is always positive,
and a force given by $F_{x}=-kx$ (Hooke's law), where k is the force
constant. Higher is the distance from the equilibrium point ($x=0$)
higher are potential energy and force, the last directed towards the
equilibrium point where both, potential and force, vanish. In the
quantum formulation there is a non-zero ground state energy (zero-point
energy).

Similarly to the harmonic oscillator, the potential (\ref{eq:V(x) integral form})
is constructed starting from a quantity, the mutual information $I_{k}(t_{k},\mathbf{x})$,
strictly positive. From the discussion we have done on the work of
Globerson et al. at the beginning of this section, the optimal (network)
system, that makes the mapping closest to the true relationship between
inputs and targets, is the one with the minimum mutual information,
which implies the elimination of the structures that are not responsible
for the true relationship contained in the measured data. Thus, in
the hypotheses of our model, given two networks consisting with observations,
the one with the largest error in the target prediction has a greater
potential energy (mutual information), being subjected to a higher
mean force. Note that for a dataset that contains some unknown relationship
between inputs and targets the potential (\ref{eq:V(x) integral form})
cannot be zero, which would imply $p(t_{k}|\mathbf{x})=p(t_{k})$
and for the joint probability $p(t_{k},\mathbf{x})=p(t_{k})p(\mathbf{x})$,
with the absence of a relation (independence) between $\mathbf{x}$
and $\mathbf{t}$ and therefore the impossibility to make a prediction.
Therefore, an expected result for the energy obtained from the application
of the potential (\ref{eq:V(x) integral form}) to the differential
equation (\ref{eq:State equation 1}) is a non-zero value for the
zero-point energy, that is an energy $E>0$ for the system at the
minimum, and a potential not null.

Considering that the network provides an approximation to the target
$t_{k}$ given by a deterministic function $y_{k}(\mathbf{x};\Gamma)$
with a noise $\varepsilon_{k}$, $t_{k}=y_{k}+\varepsilon_{k}$, and
considering that the error $\varepsilon_{k}$ is normally distributed
with mean zero, the conditional probability $p(t_{k}|\mathbf{x})$
can be written as \citep{Bishop:NNPR:1995}

\begin{equation}
p(t_{k}|\mathbf{x})=\frac{1}{(2\pi\chi_{k}^{2})^{1/2}}\exp\left\{ -\frac{(y_{k}-t_{k})^{2}}{2\chi_{k}^{2}}\right\} \label{eq:p(tk|x)}
\end{equation}
We can interpret the standard deviation $\chi_{k}$ of the predictive
distribution for $t_{k}$ as an error bar on the mean value $y_{k}$.
Note that $\chi_{k}$ depends on $\mathbf{x}$, $\chi_{k}=\chi_{k}(\mathbf{x})$,
so $\chi_{k}$ is not a variational parameter ($\chi_{k}\notin\Gamma$).
To be able to integrate the differential equation (\ref{eq:State equation 1})
we will consider the vector $\vec{\chi}$ constant. We will see at
the end of the discussion that it is possible to obtain an expression
for $\chi_{k}$ dependent on $\mathbf{x}$, which allows us to derive
a more precise description of the state function.

Writing unconditional probabilities for inputs and targets as Gaussians,
we have
\[
p(\mathbf{x})=\frac{1}{(2\pi)^{N/2}\left|\Sigma\right|^{1/2}}\exp\left\{ -\frac{1}{2}(\mathbf{x}-\vec{\mu})^{T}\sum\,^{-1}(\mathbf{x}-\vec{\mu})\right\} 
\]
\begin{equation}
p(t_{k})=\frac{1}{(2\pi\theta_{k}^{2})^{1/2}}\exp\left\{ -\frac{(t_{k}-\rho_{k})^{2}}{2\theta_{k}^{2}}\right\} \label{eq:p(tk)}
\end{equation}
Considering the absence of correlation between the $N$ input variables,
the probability $p(\mathbf{x})$ is reduced to
\begin{equation}
p(\mathbf{x})=\prod_{i=1}^{N}\frac{1}{\sqrt{2\pi}\sigma_{i}}\exp\left\{ -\frac{(x_{i}-\mu_{i})^{2}}{2\sigma_{i}^{2}}\right\} =\prod_{i=1}^{N}\mathcal{N}_{i}\label{eq:p(x) uncorrelated}
\end{equation}
where, in this case, $\left|\Sigma\right|^{1/2}=\prod_{i=1}^{N}\sigma_{i}$,
representing with $\mathcal{N}_{i}=\mathcal{N}(\mu_{i};\sigma_{i}^{2})$
the normal distribution with mean $\mu_{i}$ and variance $\sigma_{i}^{2}$
relative to the component $x_{i}$ of the vector $\mathbf{x}$. The
equations (\ref{eq:p(tk)})  and (\ref{eq:p(x) uncorrelated}) introduce
in the model a statistical description of the problem starting from
the observed data, through the set of constants $\vec{\rho}$, $\vec{\theta}$,
$\vec{\mu}$  and $\vec{\sigma}$. The integration of the equation
(\ref{eq:V(x) integral form}) over $t_{k}$ gives
\begin{equation}
V(\mathbf{x})=\prod_{i=1}^{N}\mathcal{N}_{i}\sum_{k=1}^{C}\left(\alpha_{k}y_{k}^{2}+\beta_{k}y_{k}+\gamma_{k}\right)\label{eq:V(x)}
\end{equation}
where
\begin{equation}
\alpha_{k}=\frac{1}{2\theta_{k}^{2}},\;\beta_{k}=-\frac{\rho_{k}}{\theta_{k}^{2}},\ \gamma_{k}=\frac{\rho_{k}^{2}+\chi_{k}^{2}}{2\theta_{k}^{2}}-\ln\frac{\chi_{k}\sqrt{e}}{\theta_{k}}\label{eq:alpha=00005Bk=00005D}
\end{equation}

It is known that a linear combination of Gaussians can approximate
an arbitrary function. Using a base of dimension $P$ we can write
the following expression for $y_{k}(\mathbf{x};\Gamma)$
\begin{equation}
y_{k}(\mathbf{x};\Gamma)=\sum_{p=1}^{P}w_{kp}\phi_{p}(\mathbf{x})+w_{k0}\label{eq:yk}
\end{equation}
where
\begin{equation}
\phi_{p}(\mathbf{x})=\exp\left\{ -\xi_{p}\left\Vert \mathbf{x}-\vec{\omega}_{p}\right\Vert ^{2}\right\} =\prod_{i=1}^{N}\exp\left\{ -\xi_{p}(x_{i}-\omega_{pi})^{2}\right\} \label{eq:RBF function}
\end{equation}
and where $w_{k0}$ is the bias term for the output unit $k$. The
equations (\ref{eq:yk})  and (\ref{eq:RBF function}) propose a model
of neural network of type RBF (Radial Basis Function), which contain
a single hidden layer.

Taking into account the equations (\ref{eq:Force}), (\ref{eq:V(x)})
 and (\ref{eq:yk}) the components of the force, $F_{i}$, are given
by
\begin{equation}
\begin{array}{rcl}
F_{i} & = & \zeta\prod_{i=1}^{N}\mathcal{N}_{i}\sum_{k=1}^{C}\left\{ \frac{x_{i}-\mu_{i}}{\sigma_{i}^{2}}\left(\alpha_{k}w_{k0}^{2}+\beta w_{k0}+\gamma_{k}\right)+\right.\\
 &  & \left(2\alpha_{k}w_{k0}+\beta_{k}\right)\sum_{p=1}^{P}w_{kp}\left[2\xi_{p}\left(x_{i}-\omega_{pi}\right)+\frac{x_{i}-\mu_{i}}{\sigma_{i}^{2}}\right]\phi_{p}+\\
 &  & \left.\alpha_{k}\sum_{p=1}^{P}\sum_{q=1}^{P}w_{kp}w_{kq}\left[2\xi_{q}\left(x_{i}-\omega_{qi}\right)+2\xi_{p}\left(x_{i}-\omega_{pi}\right)+\frac{x_{i}-\mu_{i}}{\sigma_{i}^{2}}\right]\phi_{p}\phi_{q}\right\} 
\end{array}\label{eq:Forza}
\end{equation}
with an expected value for the force given by the Ehrenfest theorem
\[
\left\langle \mathbf{F}\right\rangle =-\left\langle \frac{\partial V(\mathbf{x)}}{\partial\mathbf{x}}\right\rangle =-\int\Psi^{*}(\mathbf{x})\frac{\partial V(\mathbf{x)}}{\partial\mathbf{x}}\Psi(\mathbf{x})\,d\mathbf{x}
\]

\section{\label{sec:The-state-equation}The state equation}

A dimensional analysis of the potential (\ref{eq:V(x)}) shows that
the term $\alpha_{k}y_{k}^{2}-\beta_{k}y_{k}+\gamma_{k}$ is dimensionless,
and therefore its units are determined by the factor $\left|\Sigma\right|^{-1/2}$.
Since $V(\mathbf{x})$ has been obtained from mutual information,
which unit is nat if it is expressed using natural logarithms, we
will call the units of $\left|\Sigma\right|^{-1/2}$ \emph{energy
nats} or \emph{enats}.\footnote{The concrete units of factor $\left|\Sigma\right|^{-1/2}$ are dependent
on the problem. The definition of enat allows to generalize the results.} To maintain the dimensional coherence in the equation (\ref{eq:State equation 1})
we define $\epsilon=\frac{\sigma_{\mathbf{x}}^{2}}{(2\pi)^{N/2}\left|\Sigma\right|^{1/2}}$,\footnote{This setting makes it possible to incorporate $\left|\Sigma\right|^{-1/2}$
into the value of $E$, but in the continuation we will leave it explicitly
indicated.

Calculations show that the ratio between kinetic and potential energy
is very large close to the optimum. The factor $(2\pi)^{-N/2}$ in
the expression of $\epsilon$ tries to reduce this value in order
to increase the contribution of the potential to the total energy.
This choice is arbitrary and has no significant influence in the optimization
process, but only in the numerical value of $E$.} where
\[
\sigma_{\mathbf{x}}^{2}\nabla^{2}=\sum_{i=1}^{N}\sigma_{i}^{2}\frac{\partial^{2}}{\partial x_{i}^{2}}
\]
$\sigma_{\mathbf{x}}^{2}$ cannot be a constant factor independent
of the single components of $\mathbf{x}$ since in general every $x_{i}$
has its own units and its own variance. The resulting Hamiltonian
operator
\[
\hat{H}=\hat{T}+\hat{V}=-\frac{\sigma_{\mathbf{x}}^{2}}{(2\pi)^{N/2}\left|\Sigma\right|^{1/2}}\nabla^{2}+\prod_{i=1}^{N}\mathcal{N}_{i}\sum_{k=1}^{C}\left(\alpha_{k}y_{k}^{2}+\beta_{k}y_{k}+\gamma_{k}\right)
\]
is real, linear and hermitian. Hermiticity stems from the condition
that the average value of energy is a real value, $\left\langle E\right\rangle =\left\langle E^{*}\right\rangle $.\footnote{In this article we only use real functions, so the hermiticity condition
is reduced to the symmetry of the $\mathbf{H}$ and $\mathbf{S}$
matrices.} $\hat{T}$ and $\hat{V}$ represent the operators related respectively
to the kinetic and potential components of $\hat{H}$
\[
\hat{T}=-\frac{\sigma_{\mathbf{x}}^{2}}{(2\pi)^{N/2}\left|\Sigma\right|^{1/2}}\nabla^{2}
\]
\[
\hat{V}=\prod_{i=1}^{N}\mathcal{N}_{i}\sum_{k=1}^{C}\left(\alpha_{k}y_{k}^{2}+\beta_{k}y_{k}+\gamma_{k}\right)
\]

The final state equation is
\begin{equation}
E\Psi=-\frac{\sigma_{\mathbf{x}}^{2}}{(2\pi)^{N/2}\left|\Sigma\right|^{1/2}}\nabla^{2}\Psi+\prod_{i=1}^{N}\mathcal{N}_{i}\sum_{k=1}^{C}\left(\alpha_{k}y_{k}^{2}+\beta_{k}y_{k}+\gamma_{k}\right)\Psi=\left\langle T\right\rangle +\left\langle V\right\rangle \label{eq:Final state equation}
\end{equation}
where the $\left\langle T\right\rangle $ and $\left\langle V\right\rangle $
components of the total energy $E$ are the expected values of the
kinetic and potential energy respectively. Wanting to make an analogy
with wave mechanics, we can say that the equation (\ref{eq:Final state equation})
describes the motion of a particle of mass $m=\frac{(2\pi)^{N/2}\left|\Sigma\right|^{1/2}}{2}$
subject to the potential (\ref{eq:V(x)}). $\sigma_{\mathbf{x}}^{2}$,
as happens in quantum mechanics with the Planck constant, has the
role of a scale factor: the phenomenon described by the equation (\ref{eq:Final state equation})
is relevant in the range of variance for each single component $x_{i}$
of $\mathbf{x}$.

Initially, in the initial phase of this work and in the preliminary
tests, we considered an integer factor greater than 1 in the expression
of the potential, in the form $V(\mathbf{x})=\zeta\sum_{k=1}^{C}\int I_{k}(t_{k},\mathbf{x})\,dt_{k}$.
The reason was that at the minimum of energy the potential energy
is in general very small compared to kinetic energy, and the expected
value $\left\langle V\right\rangle $ could have little influence
in the final result. However, this hypothesis proved to be unfounded
for two reasons:

1. it is true that, at $\min\left\{ E\right\} $, it occurs $\frac{\left\langle T\right\rangle }{\left\langle V\right\rangle }\gg1$,
but the search for this value with the genetic algorithm illustrated
in Section \ref{sec:Results} demonstrated how $\left\langle V\right\rangle $
is determinant far from the minimum of energy, thus having a fundamental
effect in the definition of the energy state of the system;

2. the calculations show how, even in the case of considering $\zeta\gg1$,
the ratio $\frac{\left\langle T\right\rangle }{\left\langle V\right\rangle }$
remains substantially unchanged, and the overall effect of $\zeta$
in this case is to allow expected values for the mutual information
smaller than those obtained for $\zeta=1$. This fact can have a negative
effect since it can lead to a minimum of energy where the MI does
not correspond to that energy state of the system that is identified
with the true relationship that exists between $\mathbf{x}$ and $\mathbf{t}$,
and which allows the elimination of the irrelevant superstructures
in data, as we have already discussed in Section \ref{sec:The-potential}.

We have discussed the role of the operator $\hat{V}$: its variation
in the space $\mathbf{x}$ implies a force that is directed towards
the target where $V(\mathbf{x})$ and $\mathbf{F}$ are minimal. The
operator $\hat{T}$ contains the divergence of a gradient in the space
$\mathbf{x}$ and represents the divergence of a flow, being a measure
of the deviation of the state function at a point respect to the average
of the surrounding points. The role of $\nabla^{2}$ in the equation
(\ref{eq:Final state equation}) is to introduce information about
the curvature of $\Psi$. In neural networks theory a similar role
is found in the use of the Hessian matrix of the error function, calculated
in the space of weights, in conventional second order optimization
techniques.

We will assume a base of dimension $D$ for the trial function
\begin{equation}
\Psi(\mathbf{x})=\sum_{d=1}^{D}c_{d}\psi_{d}\label{eq:Psi linear combination}
\end{equation}
with the basis functions developed as a multidimensional Gaussian
\begin{equation}
\psi_{d}(\mathbf{x})=\prod_{i=1}^{N}\exp\left\{ -\lambda_{d}(x_{i}-\eta_{id})^{2}\right\} \label{eq:psi(d) base function}
\end{equation}
The $\psi_{d}$ functions are well-bahaved because they vanish at
the infinity and therefore satisfy the boundary conditions of the
problem. As we will see in Section \ref{subsec:Interpretation-of-the state function},
$\Psi(\mathbf{x})$ can be related to a probability density. The general
form of the basis functions (\ref{eq:psi(d) base function}) ultimately
allows the description of this density as a superposition of Gaussians.

From a point of view of wave mechanics, the justification of the equation
(\ref{eq:psi(d) base function}) can be found in its similarity to
the exponential part of a Gaussian Type Orbital (GTO).\footnote{There are two definitions of GTO that lead to equivalent results:
cartesian gaussian type orbital and spherical gaussian type orbital.
The general form of a cartesian gaussian type orbital is given by
\begin{equation}
G_{ijk}^{\alpha\mathbf{R}}=N_{ijk}^{\alpha}(x-R_{1})^{i}(y-R_{2})^{j}(z-R_{3})^{k}\exp\left\{ -\alpha\left|\mathbf{r}-\mathbf{R}\right|^{2}\right\} \label{eq:GTO}
\end{equation}
where $N_{ijk}^{\alpha}$ is a normalization constant. A spherical
gaussian type orbital is instead given in function of spherical harmonics,
which arise from the central nature of the force in the atom and contain
the angular dependence of the wave function. The product of GTOs in
the formalism of quantum mechanics, as it also happens in the equations
that result from the use of the equation (\ref{eq:psi(d) base function})
in the model proposed in this paper, leads to the calculation of multi-centric
integrals.

A proposal of generalization of the equation (\ref{eq:psi(d) base function}),
closer to (\ref{eq:GTO}), can be

\[
\psi_{d}(\mathbf{x})=N\prod_{i=1}^{N}(x_{i}-\eta_{id})^{\nu_{i}}\exp\left\{ -\lambda_{d}(x_{i}-\eta_{id})^{2}\right\} 
\]
where $\nu_{i}$ are variational exponents.} The difference of (\ref{eq:psi(d) base function}) respect to GTOs
simplify the integrals calculation. We can interpret $\lambda_{d}$
and $\eta_{id}$ as quantities having a similar role, respectively,
to the orbital exponent and the center in the GTOs. In some theoretical
frameworks of artificial neural networks equations (\ref{eq:Psi linear combination})
and (\ref{eq:psi(d) base function}) explicit the so called Gaussian
mixture model.

Using the equation (\ref{eq:Psi linear combination}), the expected
values for energy and for kinetic and potential terms can be written
as

\begin{equation}
\left\langle T\right\rangle =-\frac{1}{(2\pi)^{N/2}\left|\Sigma\right|^{1/2}}\int\Psi^{*}\sigma_{\mathbf{x}}^{2}\nabla^{2}\Psi\,d\mathbf{x}=-\frac{1}{\left|\Sigma\right|^{1/2}}\sum_{i=1}^{N}\sigma_{i}^{2}\sum_{m=1}^{D}\sum_{n=1}^{D}c_{m}c_{n}T_{imn}\label{eq:<T>}
\end{equation}
\begin{equation}
\left\langle V\right\rangle =\int\Psi^{*}\hat{V}\Psi\,d\mathbf{x}=\sum_{k=1}^{C}\sum_{m=1}^{D}\sum_{n=1}^{D}\sum_{k=1}^{C}c_{m}c_{n}V_{kmn}\label{eq:<V>}
\end{equation}
where
\[
T_{imn}=\int\psi_{m}^{*}\frac{\partial\psi_{n}}{\partial x_{i}^{2}}\,d\mathbf{x}
\]
\[
V_{kmn}=\int\psi_{m}^{*}\left(\alpha_{k}y_{k}^{2}+\beta_{k}y_{k}+\gamma_{k}\right)\prod_{i=1}^{N}\mathcal{N}_{i}\psi_{n}\,d\mathbf{x}
\]

Starting from the expected energy value obtained from the equation
(\ref{eq:Final state equation})\footnote{We will denote the expected value of energy, $\left\langle E\right\rangle $,
simply as $E$. Although all the functions used in this work are real,
we will make their complex conjugates explicit in the equations, as
is usual in the wave mechanics formulation.}

\begin{equation}
E=\frac{\int\Psi{}^{*}\hat{H}\Psi\,d\mathbf{x}}{\int\Psi{}^{*}\Psi\,d\mathbf{x}}\label{eq:Energy expected value}
\end{equation}
and considering the coefficients in equation (\ref{eq:Psi linear combination})
independent of each other, $\frac{\partial c_{m}}{\partial c_{n}}=\delta_{mn}$,
where $\delta_{mn}$ is the Kronecker delta, the Rayleigh-Ritz method
leads to the linear system
\begin{equation}
\sum_{n}\left[\left(H_{mn}-S_{mn}E\right)c_{n}\right]=0\label{eq:Secular system}
\end{equation}
where
\begin{equation}
H_{mn}=\int\psi_{m}^{*}\hat{H}\psi_{n}\,d\mathbf{x}\label{eq:Hmn}
\end{equation}
\begin{equation}
S_{mn}=\int\psi_{m}^{*}\psi_{n}\,d\mathbf{x}\label{eq:Smn}
\end{equation}
Condition $S_{mn}=\delta_{mn}$ cannot be assumed due to the, in general,
non-orthonormality of the basis (\ref{eq:psi(d) base function}).
Orthonormal functions can be obtained, for example, with the Gram-Schmidt
method or using a set of functions of some hermitian operator.

To obtain a nontrivial solution the determinant of the coefficients
have to be zero
\begin{equation}
\det\left(\mathbf{H}-\mathbf{S}E\right)=0\label{eq:Secular determinant}
\end{equation}
which leads to $D$ energies, equal to the size of the base (\ref{eq:Psi linear combination}).
The $D$ energy values $E_{d}$ represent an upper limit to the first
true energies $E_{d}^{'}$ of the system. The substitution of every
$E_{d}$ in (\ref{eq:Secular system}) allows to calculate the $D$
coefficients $c$ of $\Psi$ relative to the state $d$. The lowest
value among $E_{d}$ represents the global optimum of the problem
or \emph{fundamental state} that leads, in the hypotheses of this
paper, to the minimum or global error of the neural network in the
prediction of the target $\mathbf{t}$, while the remaining eigenvalues
can be interpreted as local minima. It can be shown that the eigenfunctions
obtained in this way form an orthogonal set. The variational method
we have discussed has a general character and can be applied, in principle,
to artificial neural networks of any kind, not bound to any specific
functional form for $y_{k}$.

Using equations (\ref{eq:yk}) and (\ref{eq:RBF function}) and taking
into account the constancy of $\vec{\chi}$, the integrals (\ref{eq:Hmn})
and (\ref{eq:Smn}) have the following expressions
\[
S_{mn}=\left(\frac{\pi}{{\it \lambda_{n}}+{\it \lambda_{m}}}\right)^{\frac{N}{2}}\prod_{i=1}^{N}\exp\left\{ -\frac{\lambda_{m}\lambda_{n}}{{\it \lambda_{n}}+{\it \lambda_{m}}}(\eta_{im}-\eta_{in})^{2}\right\} 
\]
\[
\begin{array}{rcl}
H_{mn} & = & -\frac{2}{\left|\Sigma\right|^{1/2}}\frac{\lambda_{m}\lambda_{n}}{\lambda_{n}+\lambda_{m}}S_{mn}\times\\
 &  & \sum_{i=1}^{N}\sigma_{i}^{2}\left[2\frac{\lambda_{m}\lambda_{n}}{{\it \lambda_{n}}+{\it \lambda_{m}}}(\eta_{im}-\eta_{in})^{2}-1\right]+\\
 &  & \zeta\left(\Lambda_{mn}\sum_{k=1}^{C}\gamma_{k}+\Lambda_{mn}\sum_{k=1}^{C}\beta_{k}w_{k0}+\right.\\
 &  & \sum_{k=1}^{C}\beta_{k}\sum_{p=1}^{P}w_{kp}\Omega_{mnp}+\Lambda_{mn}\sum_{k=1}^{C}\alpha_{k}w_{k0}^{2}+\\
 &  & 2\sum_{k=1}^{C}\alpha_{k}w_{k0}\sum_{p=1}^{P}w_{kp}\Omega_{mnp}+\\
 &  & \left.\sum_{k=1}^{C}\alpha_{k}\sum_{p=1}^{P}\sum_{q=1}^{P}w_{kp}w_{kq}\Phi_{mnpq}\right)
\end{array}
\]
where
\[
\begin{array}{rcl}
\Lambda_{mn} & = & \prod_{i=1}^{N}\frac{1}{\sqrt{2\sigma_{i}^{2}\left(\lambda_{n}+\lambda_{m}\right)+1}}\times\\
 &  & \exp\left\{ -\frac{2\sigma_{i}^{2}\left(\eta_{in}-\eta_{im}\right)^{2}\lambda_{m}\lambda_{n}+\left(\eta_{in}-\mu_{i}\right)^{2}\lambda_{n}+\left(\eta_{im}-\mu_{i}\right)^{2}\lambda_{m}}{2\sigma_{i}^{2}\left(\lambda_{n}+\lambda_{m}\right)+1}\right\} 
\end{array}
\]
\[
\begin{array}{rcl}
\Omega_{mnp} & = & \prod_{i=1}^{N}\left[\frac{1}{\sqrt{2\sigma_{i}^{2}\left(\xi_{p}+\lambda_{n}+\lambda_{m}\right)+1}}\times\right.\\
 &  & \exp\left\{ \frac{2\left(2\sigma_{i}^{2}\left(\eta_{in}\lambda_{n}+\eta_{im}\lambda_{m}\right)+\mu_{i}\right)\xi_{p}\omega_{pi}}{2\sigma_{i}^{2}\left(\xi_{p}+\lambda_{n}+\lambda_{m}\right)+1}\right\} \times\\
 &  & \exp\left\{ -\frac{\left(2\sigma_{i}^{2}\left(\lambda_{n}+\lambda_{m}\right)+1\right)\xi_{p}\omega_{pi}^{2}+\left(2\sigma_{i}^{2}\left(\eta_{in}^{2}\lambda_{n}+\eta_{im}^{2}\lambda_{m}\right)+\mu_{i}^{2}\right)\xi_{p}}{2\sigma_{i}^{2}\left(\xi_{p}+\lambda_{n}+\lambda_{m}\right)+1}\right\} \times\\
 &  & \left.\exp\left\{ -\frac{2\sigma_{i}^{2}\left(\eta_{in}-\eta_{im}\right)^{2}\lambda_{n}\lambda_{m}+\left(\eta_{in}-\mu_{i}\right)^{2}\lambda_{n}+\left(\eta_{im}-\mu_{i}\right)^{2}\lambda_{m}}{2\sigma_{i}^{2}\left(\xi_{p}+\lambda_{n}+\lambda_{m}\right)+1}\right\} \right]
\end{array}
\]
\[
\begin{array}{rcl}
\Phi_{mnpq} & = & \prod_{i=1}^{N}\left[\frac{1}{\sqrt{2\sigma_{i}^{2}\left(\xi_{q}+\xi_{p}+\lambda_{n}+\lambda_{m}\right)+1}}\times\right.\\
 &  & \exp\left\{ \frac{2\left(2\sigma_{i}^{2}\left(\xi_{p}\omega_{pi}+\eta_{in}\lambda_{n}+\eta_{im}\lambda_{m}\right)+\mu_{i}\right)\xi_{q}\omega_{qi}-2\sigma_{i}^{2}\left(\eta_{in}-\eta_{im}\right)^{2}\lambda_{n}\lambda_{m}}{2\sigma_{i}^{2}\left(\xi_{q}+\xi_{p}+\lambda_{n}+\lambda_{m}\right)+1}\right\} \times\\
 &  & \exp\left\{ -\frac{\left(2\sigma_{i}^{2}\left(\xi_{p}+\lambda_{n}+\lambda_{m}\right)+1\right)\xi_{q}\omega_{qi}^{2}+\left(2\sigma_{i}^{2}\left(\xi_{p}\omega_{pi}^{2}+\eta_{in}^{2}\lambda_{n}+\eta_{im}^{2}\lambda_{m}\right)+\mu_{i}^{2}\right)\xi_{q}}{2\sigma_{i}^{2}\left(\xi_{q}+\xi_{p}+\lambda_{n}+\lambda_{m}\right)+1}\right\} \times\\
 &  & \exp\left\{ -\frac{\left(2\sigma_{i}^{2}\left(\lambda_{n}+\lambda_{m}\right)+1\right)\xi_{p}\omega_{pi}^{2}-2\left(2\sigma_{i}^{2}\left(\eta_{in}\lambda_{n}+\eta_{im}\lambda_{m}\right)+\mu_{i}\right)\xi_{p}\omega_{pi}}{2\sigma_{i}^{2}\left(\xi_{q}+\xi_{p}+\lambda_{n}+\lambda_{m}\right)+1}\right\} \times\\
 &  & \left.\exp\left\{ -\frac{\left(2\sigma_{i}^{2}\left(\eta_{in}^{2}\lambda_{n}+\eta_{im}^{2}\lambda_{m}\right)+\mu_{i}^{2}\right)\xi_{p}+\left(\eta_{in}-\mu_{i}\right)^{2}\lambda_{n}+\left(\eta_{im}-\mu_{i}\right)^{2}\lambda_{m}}{2\sigma_{i}^{2}\left(\xi_{q}+\xi_{p}+\lambda_{n}+\lambda_{m}\right)+1}\right\} \right]
\end{array}
\]
The number of variational parameters of the model, $n_{\Gamma}$,
is
\begin{equation}
n_{\Gamma}=C(P+1)+(N+1)P+(D+2)N+D+2C+1\label{eq:Number variational parameters}
\end{equation}

The energies obtained by the determinant (\ref{eq:Secular determinant})
allow to obtain a system of equations resulting from the condition
of minimum
\begin{equation}
\frac{\partial E_{d}}{\partial\Gamma}=0\label{eq:Equations system on parameters}
\end{equation}
The system (\ref{eq:Equations system on parameters}) is implicit
in $\chi_{k}$ and must be solved in an iterative way, as $\chi_{k}$
depends on $y_{k}$ which in turn is a function of $\Gamma=\Gamma(\chi_{k})$.

One of the strengths of the proposed model is the potential possibility
of allowing the application quantum mechanics results to the study
of neural networks. An example is constituted of the generalized Hellmann-Feynman
theorem
\[
\frac{\partial E_{d}}{\partial\Gamma}=\int\Psi^{*}\frac{\partial\hat{H}}{\partial\Gamma}\Psi\,d\mathbf{x}
\]
whose validity needs to be demonstrated in this context, but whose
use seems justified since it can be demonstrated by assuming exclusively
the normality of $\Psi$ and the hermiticity of $\hat{H}$. Its applicability
could help in the calculation of the system (\ref{eq:Equations system on parameters}).

\section{\label{sec:System-and-dataset}System and dataset}

The model proposed in the previous sections implements a procedure
that realizes the optimization of a neural network for a given problem.
This optimization is subject to the minimization of the value of a
physical property of the system, the energy, calculated from the equation
(\ref{eq:Final state equation}). The potential energy term can be
considered a measure of the distance between the probability distributions
of inputs and targets \citep{steuer_measuring_2005,villaverde_mider_2014},
given by the mutual information.\footnote{The properties of mutual information, such as symmetry, allow us to
consider it as a distance relatively to the calculations made in this
paper.} More precisely, the potential energy is the expected value of the
mutual information, that is the mean value of MI taking into account
all possible values of $\mathbf{x}$ and $\mathbf{t}$.

The dataset contains the input and output measurements. We can consider
it composed of two matrices, respectively $\mathbf{x}$ and $\mathbf{t}$,
with the following structure
\[
\mathbf{x=}\left(\begin{array}{cccc}
x_{11} & x_{12} & \cdots & x_{1N}\\
x_{21} & x_{22} & \cdots & x_{2N}\\
\vdots & \vdots & \ddots & \vdots\\
x_{\mathcal{D}1} & x_{\mathcal{D}2} & \cdots & x_{\mathcal{D}N}
\end{array}\right),\,\mathbf{t}=\left(\begin{array}{cccc}
t_{11} & t_{12} & \cdots & t_{1C}\\
t_{21} & t_{22} & \cdots & t_{2C}\\
\vdots & \vdots & \ddots & \vdots\\
t_{\mathcal{D}1} & t_{\mathcal{D}2} & \cdots & t_{\mathcal{D}C}
\end{array}\right)
\]
where $\mathcal{D}$\nomenclature[D]{$\mathcal{D}$}{number of record in the dataset}
represents the number of records in the dataset. The interpretation
of these matrices is as follows:
\begin{enumerate}
\item Matrix $\mathbf{x}$ is composed of $N$ vectors $x_{i}$, equal to
the number of columns and which correspond to the number of features
of the problem. Vectors $x_{i}$ are represented in italic and not
in bold as they represent independent variables in the formalism of
the EANNs. A similar discourse can apply to the vectors $t_{k}$.
\item Matrix $\mathbf{t}$ is composed of $C$ vectors $t_{k}$, equal to
the number of columns and which correspond to the number of output
units of the network.
\item Each $x_{ji}$ represents the $j$th record value for feature $i$.
\item Each $t_{jk}$ represents the value of the $j$th record relative
to the output $k$.
\end{enumerate}
Table \ref{tab:Equivalence table} illustrates the equivalence between
the EANN formalism and quantum mechanics. In particular:
\begin{enumerate}
\item Schrödinger's equation describes the motion of the electron in the
space subject to a force. Mathematically, the number of independent
variables is therefore the number of spatial coordinates. In an EANN
an equivalent system would be represented by a problem with $N=3$.
This constitutes a substantial difference between the two models,
since in an EANN the value of $N$ is not fixed but problem-dependent.
\item The different electron positions in a quantum system are equivalent
to the individual measurements $x_{ji}$ once the target is observed.
\item The probability of finding the electron in a region of space is ultimately
defined by the wave function. Similarly, in an EANN the probability
of making a measurement within a region in the feature space after
observing the target is related to the system state function, as we
will analyze in the Section \ref{subsec:Interpretation-of-the state function}.
\item Both models depend on a series of constant characteristics of the
system: electron charge and Planck constant in the Schrõdinger equation,
$\vec{\rho}$, $\vec{\theta}$, $\vec{\mu}$, $\vec{\sigma}$ e $\mathbf{w}$
in an EANN.
\end{enumerate}
\begin{table}
\caption{\label{tab:Equivalence table}Equivalence between the formalism of
EANNs and wave mechanics.}

\centering{}%
\begin{tabular}{lcl}
\hline 
Property & EANN & Schrõdinger equation\tabularnewline
\hline 
\hline 
Input size & $N$ & 3\tabularnewline
Output size & $C$ & \tabularnewline
Features & $x_{i}$ & Spatial coordinates\tabularnewline
Input measurements & $x_{ji}$ & Electron position\tabularnewline
System constants & $\vec{\rho}$, $\vec{\theta}$, $\vec{\mu}$, $\vec{\sigma}$, $\mathbf{w}$ & Electron charge ($e$)\tabularnewline
 &  & Planck constant ($h)$\tabularnewline
\hline 
\end{tabular}
\end{table}

\section{\label{sec:Interpretation}Interpretation of the model}

\subsection{\label{subsec:Interpretation-of-the state function}Interpretation
of the state function}

The model we have proposed contains three main weaknesses: 1) the
normality of the marginal densities $p(\mathbf{x})$ and $p(\mathbf{t})$;
2) the absence of correlation between the $N$ components of the input
$\mathbf{x}$; 3) the constancy of the vector $\vec{\chi}$. The following
discussion tries to analyze the third point.

Similarly to wave mechanics an the Born rule, we can interpret $\Psi$
as a probability amplitude and the square module of $\Psi$ as a probability
density. In this case, the Laplacian operator in equation (\ref{eq:Final state equation})
models a probability flow. Given that we have obtained $\Psi$ from
a statistical description of a set of known targets, we can assume
that $\left|\Psi\right|^{2}$ represents the conditional probability
of $\mathbf{x}$ given $\mathbf{t}$ subject to the set of parameters
$\Gamma$

\begin{equation}
p(\mathbf{x}|\mathbf{t},\Gamma)=\left|\Psi(\mathbf{x})\right|^{2}\label{eq:p(x|t)=00003D|Psi|^2}
\end{equation}
where $\left|\Psi\right|^{2}d\mathbf{x}$ represents the probability,
given $\mathbf{t}$, of finding the input between $\mathbf{x}$ and
$\mathbf{x}+d\mathbf{x}$. In this interpretation $\Psi$ is a conditional
probability amplitude.

A fundamental aspect of this paper is that this interpretation of
$\left|\Psi(\mathbf{x})\right|^{2}$ as conditional probability is
to be understood in a classical statistical sense, which allows to
connect the quantum probabilities of the formalism of the EANNs with
fundamental statistical results in neural networks and Bayesian statistics,
such as Bayes' theorem. This view is in agreement with the work of
some authors in quantum physics, who have underlined how the interpretation
of Born's rule as a classical conditional probability is the connection
that links quantum probabilities with experience \citep{hofer-szabo_quantum_2019}.\footnote{The conditional nature of quantum probability also appears in other
contexts. In the formalism of Bohmian Mechanics, consider, for example,
a particular particle with degree of freedom $x$. We can divide the
generic configuration point $q=\left\{ x,y\right\} $, where $y$
denotes the coordinates of other particles, i.e., degrees of freedom
from outside the subsystem of interest. We then define the conditional
wave function (CWF) for the particle as follows:
\[
\phi(x,t)=\left.\Psi(q,t)\right|{}_{y=Y(t)}=\Psi\left[x,Y(t),t\right]
\]
That is, the CWF for one particle is simply the universal wave function,
evaluated at the actual positions $Y(t)$ of the other particles \citep{norsen_bohmian_2016}.

This formalization tells us that any a priori knowledge, not strictly
belonging to the system under study, leads us to the concept of conditional
amplitude and conditional probability density.} Given the relevance of this point of view in the mathematical treatment
that follows, we will make a hypothetical example that in our opinion
can better clarify this aspect.

Consider a system consisting of a hydrogen-like atom, with a single
electron. Suppose that this system is located in a universe, different
from our universe, in which the following conditions occur:
\begin{enumerate}
\item Schrödinger's equation is still a valid description of the system;
\item the potential has the known expression $V=-\frac{Ze^{2}}{r}$, where
$Z$ is the number of protons in the nucleus of the atom (atomic number),
$e$ the charge of the electron and $r$ the distance between proton
and electron;
\item the electron charge is constant given a system, but has different
values for different systems;
\item only a value of $e$ can be used in the Schrödinger equation;
\item $e$ is dependent on the system for which it is uniquely defined,
but is independent of spatial coordinates and time;
\item the numerical values for energy and other properties of the atom obtained
with different $e$ values are different.
\item different physicists will agree in the choice of $e$. However, a
laboratory measurement campaign is required to identify its value.
\item It is not possible to derive the value of $e$ used for a system from
the energy and state function calculated for that system.
\end{enumerate}
Imagine a physicist applying Schrödinger's equation with a given $e$
value for the electron charge, obtaining a spatial probability density
of the electron $\left|\Psi(\mathbf{x})\right|^{2}$. Since $e$ is
not unique and cannot be obtained through a logical or mathematical
reasoning but only from laboratory measurements, our physicist will
have to indicate which value for $e$ he used in his calculations,
in order to allow others scientists to repeat his work. So, he will
have to say that he got an expression for $\left|\Psi(\mathbf{x})\right|^{2}$
and $E$ given a certain value for $e$, for example writing
\[
\left|\Psi(\mathbf{x}|e)\right|^{2}
\]
This notation underlines the conditional character of the electron
spatial probability density. Since in our universe the electron charge
is constant in all physical systems and the choice of its value is
unique, we can simply write $\left|\Psi(\mathbf{x})\right|^{2}$.
In a EANNs the probability density in the input space $\left|\Psi(\mathbf{x})\right|^{2}$
is obtained from a potential that contains constants ($\alpha_{k}$,
$\beta_{k}$, $\gamma_{k}$) calculated from a conditional probability
density ($p(\mathbf{t}|\mathbf{x}))$ and marginal densities ($p(\mathbf{x}),p(\mathbf{t}))$
dependent on the target of the problem.

The character of $\left|\Psi(\mathbf{x})\right|^{2}$ as a classical
probability allows to relate the equation (\ref{eq:p(x|t)=00003D|Psi|^2})
with the conditional probability $p(\mathbf{t}|\mathbf{x})$ through
Bayes' theorem
\begin{equation}
p(\mathbf{x}|\mathbf{t})=\frac{p(\mathbf{t}|\mathbf{x})p(\mathbf{x})}{p(\mathbf{t})}\label{eq:Bayes theorem}
\end{equation}
Since we considered the $C$ targets independent, using the expressions
(\ref{eq:p(tk|x)}) and (\ref{eq:p(x|t)=00003D|Psi|^2}) into (\ref{eq:Bayes theorem}),
separating variables and integrating over $t_{k}$, assuming that
at the optimum is satisfied the condition $\theta_{k}>\chi_{k}$,
we have
\[
\frac{\left|\Psi(\mathbf{x})\right|^{2}}{p(\mathbf{x})}=\prod_{k=1}^{C}\sqrt{\frac{2\pi}{\theta_{k}^{2}-\chi_{k}^{2}}}\theta_{k}^{2}\exp\left\{ \frac{(y_{k}(\mathbf{x})-\rho_{k})^{2}}{2\left(\theta_{k}^{2}-\chi_{k}^{2}\right)}\right\} 
\]
which leads to an implicit equation in $\chi_{k}$. For networks with
a single output, $C=1$, we have
\begin{equation}
\chi_{(\tau+1)}=\sqrt{\theta^{2}-2\pi\theta^{4}\frac{p(\mathbf{x})^{2}}{\left|\Psi_{(\tau)}\right|^{4}}\exp\left\{ \frac{(y-\rho)^{2}}{\theta^{2}-\chi_{(\tau)}^{2}}\right\} }\label{eq:tau + 1}
\end{equation}
where $\Psi$, $y$ and $\chi$ are functions of $\mathbf{x}$. The
equation (\ref{eq:tau + 1}) allows in principle an iterative procedure
which, starting from the constant initial value $\chi_{(0)}$ which
leads to a state function $\Psi_{(0)}$, through the resolution of
the system (\ref{eq:Secular system}) permits to calculate successive
corrections of $\Psi$.

\subsection{Superposition of states and probability}

As we said in Section \ref{sec:The-state-equation} the basis functions
(\ref{eq:psi(d) base function}) are not necessarily orthogonal, which
forces to calculate the overlap integrals given that in this case
$S_{mn}\neq\delta_{mn}$, where $\delta_{mn}$ is the Kronecker delta.
The basis functions can be made orthogonal or a set of functions of
some hermitian operator can be chosen, which can be demonstrate to
form a complete orthogonal basis set. In the case of a function (\ref{eq:Psi linear combination})
result of the linear combination of single orthonormal states, the
coefficient $c_{m}$ represents the probability amplitude $\left\langle \psi_{m}|\Psi\right\rangle $
of state $m$
\[
\left\langle \psi_{m}|\Psi\right\rangle =\sum_{d=1}^{D}c_{d}\int\psi_{m}^{*}\psi_{d}\,d\mathbf{x}=\sum_{d=1}^{D}c_{d}\delta_{dm}=c_{m}
\]
being in this case $\sum_{d=1}^{D}\left|c_{d}\right|^{2}=1$ and $\left|c_{d}\right|^{2}$
the prior probability of the state $d$.

\subsection{Work and complexity}

From a physical point of view, the motion of a particle within a conservative
force field implies a potential difference and the associated concept
of work, done by the force field or carried out by an external force.
In physical conservative fields, work, $W$, is defined as the minus
difference between the potential energy of a body subject to the force
field and that possessed by the body at a an arbitrary reference point,
$W=-\Delta V(\mathbf{x})$. In some cases of central forces, as in
the electrostatic or gravitational ones, the reference point is located
at an infinite distance from the source where, given the dependence
of $V$ on $\frac{1}{r}$, the potential energy is zero.

Consider a neural network immersed in the field generated by the targets,
as described in the previous sections, and which we will call \emph{bounded
system}. We can consider this system as the result of the trajectory
followed by a single network with respect to a reference point as
starting point. The definition of potential given in Section \ref{sec:The-potential}
allows to calculate a potential difference between both points, which
implies physical work carried out by the force field or provided by
an external force. In the latter case, it may be possible to arrive
at a situation where the end point of the process is the unbounded
system (\emph{free system}), not subject to the influence of the field
generated by the targets. In this case, the amount of energy provided
to the system is an analogue of the ionization energy in an atom.

If we maintain the convention of signs of physics, $W>0$ means a
work done by the force on the bodies immersed in the field and a system
that evolves towards a more stable configuration with lower potential
energy. For $W<0$, however, the system evolves towards a greater
potential configuration that can only be achieved through the action
of an external force. The potential energy is one of the components
of the total energy and a decrease in the potential energy does not
necessarily imply a decrease of the total energy, but for a stable
constrained system the potential of the final state will be lower
than the potential of the initial state, represented by the reference
point, if $W>0$. From these considerations, a good reference point
should have the following two properties:
\begin{enumerate}
\item have a maximum value with respect to the potential of any system that
can be modeled by the equation (\ref{eq:Final state equation});
\item be independent of the system.
\end{enumerate}
Some well-known basic results from information theory allow to define
an upper limit for mutual information. From the general relations
that link mutual information, differential entropy, conditional differential
entropy and joint differential entropy, we have the following equivalent
expressions for a target $t_{k}$
\begin{equation}
\iint I_{k}(t_{k},\mathbf{x})\,dt_{k}d\mathbf{x}=h(t_{k})-h(t_{k}|\mathbf{x})=h(\mathbf{x})-h(\mathbf{x}|t_{k})\geq0\label{eq:I and h bis}
\end{equation}
\begin{equation}
\iint I_{k}(t_{k},\mathbf{x})\,dt_{k}d\mathbf{x}=h(t_{k})+h(\mathbf{x})-h(t_{k},\mathbf{x})\geq0\label{eq:I and h}
\end{equation}
In the case that $h(t_{k})>0,\,h(\mathbf{x})>0$ we can write the
following inequalities

\begin{equation}
\iint I_{k}(t_{k},\mathbf{x})\,dt_{k}d\mathbf{x}<\left\{ \begin{array}{l}
h(t_{k})\\
h(\mathbf{x})
\end{array}\right.\label{eq:Relation I and h}
\end{equation}
Equation (\ref{eq:Relation I and h}) allows to postulate an equivalent
condition which constitutes an upper limit for the expected value
of the potential energy given the expected value for differential
entropies

\begin{equation}
\left\langle V\right\rangle <\left\{ \begin{array}{l}
\left\langle h(t_{k})\right\rangle \\
\left\langle h(\mathbf{x})\right\rangle 
\end{array}\right.\label{eq:Relation V and h}
\end{equation}
where $\left\langle h(t_{k})\right\rangle $ and $\left\langle h(\mathbf{x})\right\rangle $
are given by
\[
\left\langle h(t_{k})\right\rangle =\iint\Psi^{*}h(t_{k})\Psi\,dt_{k}d\mathbf{x}
\]
\[
\left\langle h(\mathbf{x})\right\rangle =\int\Psi^{*}h(\mathbf{x})\Psi\,d\mathbf{x}
\]

Expressions (\ref{eq:Relation I and h}) and (\ref{eq:Relation V and h})
establish that the mutual information and ultimately the expected
value of the potential energy are upper limited by a maximum value
given by the entropy of the distribution of inputs or targets. As
we have already discussed, to choose this maximum value as an arbitrary
reference point it is desirable to identify the differential entropies
from probability densities independent of the problem under consideration.
Adequate choices can be the entropy of the uniform distribution or
the entropy of the normal distribution with the same variance of the
dataset. To analyze more deeply the physical nature of the probability
density which gives rise to the maximum entropy, we will now consider
the question from a point of view closer to quantum mechanics.

Consider a system defined by a state function $\Psi_{0}$ in which
inputs and targets are independent
\[
p(\mathbf{t},\mathbf{x})=p(\mathbf{t})p(\mathbf{\mathbf{x}})
\]
Such a system has zero mutual information. Assuming valid the interpretation
we gave in the Section \ref{subsec:Interpretation-of-the state function}
of the square of the state function as conditional density $p(\mathbf{x}|\mathbf{t})$,
together with the equation (\ref{eq:I and h bis}), calling $h_{0}$
the differential entropy of the system, we have\nomenclature[Psi0]{$\Psi_{0}$}{state function for free system}
\[
h\left(\left|\Psi_{0}\right|^{2}\right)=h_{0}(\mathbf{x})
\]
This system is not bound and we will call it \emph{free system}, in
analogy with that of the free particle in quantum mechanics. In our
model it is described by the following state equation
\[
-\frac{\sigma_{\mathbf{x}}^{2}}{(2\pi)^{N/2}\left|\Sigma\right|^{1/2}}\nabla^{2}\Psi_{0}=E\Psi_{0}
\]
The multidimensional problem can be reduced to $N$ one-dimensional
problems
\begin{equation}
\Psi_{0}(\mathbf{x})=\prod_{i=1}^{N}A_{i}\psi_{0}(x_{i})=A\prod_{i=1}^{N}\psi_{0}(x_{i})\label{eq:Psi0 free system}
\end{equation}
which gives an energy
\[
E=\sum_{i=1}^{N}E_{i}
\]
and where $A$ is the normalization constant. There are two solutions
for the one-dimensional stationary system

\begin{equation}
\psi_{0}^{\pm}(x_{i})=A_{i}^{\pm}\exp\left\{ \pm i\sqrt{\frac{(2\pi)^{1/2}E_{i}}{\sigma_{i}}}x_{i}\right\} \label{eq:Psi free system}
\end{equation}
where $E$ is not quantized and any energy satisfying $E\geq0$ is
allowed. The previous equation corresponds to two plane waves, one
moving to the right and the other to the left of the $x_{i}$ axis.
The general solution can be written as a linear combination of both
solutions.

The normalization of this system is problematic because the state
function cannot be integrated and the normalization constant can only
be obtained considering a limited interval $\Delta$, but the probability
in a differential element on this interval can be calculated. Taking
any of the solutions (\ref{eq:Psi free system})

\[
\int_{\Delta}\left|\psi_{0}(x_{i})\right|^{2}\,dx_{i}=A_{i}^{2}\int_{\Delta}\exp\left\{ \pm i\sqrt{\frac{(2\pi)^{1/2}E_{i}}{\sigma_{i}}}x_{i}\right\} \exp\left\{ \mp i\sqrt{\frac{(2\pi)^{1/2}E_{i}}{\sigma_{i}}}x_{i}\right\} \,dx_{i}=1
\]
\[
A_{i}=\frac{1}{\sqrt{\Delta}}
\]
The probability density for the differential element $dx_{i}$ is

\[
\begin{array}{rcl}
\left|\psi_{0}(x_{i})\right|^{2}\,dx_{i} & = & \frac{\psi_{0}^{*}(x_{i})\psi_{0}(x_{i})\,dx_{i}}{\int_{\Delta}\psi_{0}^{*}(x_{i})\psi_{0}(x_{i})\,dx_{i}}\\
 & = & \frac{A_{i}^{2}\exp\left\{ \pm ii\sqrt{\frac{(2\pi)^{1/2}E_{i}}{\sigma_{i}}}x_{i}\right\} \exp\left\{ \mp i\sqrt{\frac{(2\pi)^{1/2}E_{i}}{\sigma_{i}}}x_{i}\right\} \,dx_{i}}{A_{i}^{2}\int_{\Delta}\exp\left\{ \pm i\sqrt{\frac{(2\pi)^{1/2}E_{i}}{\sigma_{i}}}x_{i}\right\} \exp\left\{ \mp i\sqrt{\frac{(2\pi)^{1/2}E_{i}}{\sigma_{i}}}x_{i}\right\} \,dx_{i}}\\
 & = & \frac{dx_{i}}{\Delta}
\end{array}
\]
so the probability density is constant in all points of the interval
$\Delta$ and equal to $\frac{1}{\Delta}$, and $\left|\psi_{0}(x_{i})\right|^{2}$
is the density of the uniform distribution.\footnote{This is because the state function has no boundary conditions, that
is, it does not cancel itself in any point of the space. In quantum
mechanics, from the uncertainty principle, it is equivalent to knowing
exactly the moment and having total uncertainty about the position.} Since the total state function is the product of the single one-dimensional
functions, if the normalization interval (equal to the integration
domain that we used) is the same for all the inputs, we have

\[
\left|\Psi_{0}\right|^{2}=\frac{1}{\Delta^{N}}
\]
with a differential entropy given by
\begin{equation}
h_{0}(\mathbf{x})=h\left(\left|\Psi_{0}\right|^{2}\right)=-\int_{\Delta}\frac{1}{\Delta^{N}}\ln\left(\frac{1}{\Delta^{N}}\right)\,d\mathbf{x}=N\ln\Delta=\sum_{i=1}^{N}h_{0}(x_{i})\label{eq:hmax}
\end{equation}
Equation (\ref{eq:hmax}) expresses the maximum differential entropy
considering all possible probability distributions of systems that
are solution of the equation (\ref{eq:Final state equation}), equal
to the entropy of the conditional probability density given by the
square of the state function for the free system. This maximum value
and its derivation from a density of probability independent of any
system with non-zero potential, allow its choice as a reference value
for the calculation of the potential difference. The last equality
is a consequence of the factorization of the total state function
(\ref{eq:Psi0 free system}), and expresses the multivariate differential
entropy as the sum of the single univariate entropies, which is maximum
with respect to each joint differential entropy and indicates independence
of the variables $x_{i}$. Given the constancy of $h_{0}(\mathbf{x})$,
for a normalized state function $\Psi_{0}$ in the interval $\Delta$,
the expected value of the differential entropy for the free system
is
\[
\left\langle h_{0}(\mathbf{x})\right\rangle =N\ln\Delta
\]

Considering an initial state represented by the reference point and
a final state given by the potential calculated for a proposal of
solution $y(\mathbf{x};\Gamma)$, work is given by
\begin{equation}
W=-\Delta V=h\left(\left|\Psi_{0}\right|^{2}\right)-\left\langle V\right\rangle =N\ln\Delta-\left\langle V\right\rangle \label{eq:W}
\end{equation}
For $W>0$, equation (\ref{eq:W}) explains the work, in enats, done
by the force field to pass from a neural network that makes uniformly
distributed predictions to a network that makes an approximation to
the density $p(\mathbf{t}|\mathbf{x})$. Conversely, using a terminology
proper of atomic physics, equation (\ref{eq:W}) expresses the work
that must be done on the system in order to pass from the bounded
system to the free system, the last represented by a network that
makes uniformly distributed predictions.

Consider the following equivalent expression for equation (\ref{eq:W}),
obtainable for Gaussian and normalized probability densities
\begin{equation}
W=-\iint\Psi^{*}(\mathbf{x})p(\mathbf{t},\mathbf{x})\ln\left(\frac{\frac{p(\mathbf{t},\mathbf{x})}{p(\mathbf{t})p(\mathbf{x})}}{\Delta^{N}}\right)\Psi(\mathbf{x})\,d\mathbf{t}d\mathbf{x}\label{eq:W hmax scaling factor}
\end{equation}
In equation (\ref{eq:W hmax scaling factor}) $\Delta^{N}$ has the
role of a scaling factor, similar to the function $m(x)$ in the differential
entropy as proposed by Jaynes and Guia\c{s}u \citep{baran_multi-objective_2001,baran_extension_2017,rosenkrantz_e._1989,guiasu_information_1977}.
From this point of view, $W$ is scale invariant, provided that $\left\langle V\right\rangle $
and $h\left(\left|\Psi_{0}\right|^{2}\right)$ are measured on the
same interval, and represents a variation of information, that is,
the reduction in the amount of uncertainty in the prediction of the
target through observing input with respect to a reference level given
by $h\left(\left|\Psi_{0}\right|^{2}\right)$. In this interpretation,
where we can consider $W$ as a difference in the information content
between the initial and final states of a process, $h\left(\left|\Psi_{0}\right|^{2}\right)$
is the entropy of an a priori probability and (\ref{eq:W}) is the
definiton of self-organization \citep{feldman_statistical_1997,fernandez_information_2013,gershenson_complexity_2012,lopez-ruiz_statistical_1995,santamaria-bonfil_measuring_2016}
\[
\mathcal{S}=\mathcal{I}_{i}-\mathcal{I}_{f}
\]
where $\mathcal{I}_{i}$ is the information of the initial state and
$\mathcal{I}_{f}$ is the information of the final state, represented
by the expected value for the potential energy at the optimum. In
a normalized version of $\mathcal{S}$ we have
\begin{equation}
\mathcal{S}=1-\frac{\left\langle V\right\rangle }{N\ln\Delta}\label{eq:self-organization}
\end{equation}
This implies that self-organization occurs ($\mathcal{S}>0$) if the
process reduces information, i.e. $\mathcal{I}_{i}>\mathcal{I}_{f}$.
If the process generates more information, $S<0$, emergence occurs.
Emergence is a concept complementary to the self-organization and
is proportional to the ratio of information generated by a process
with respect the maximum information
\begin{equation}
\mathcal{E}=\frac{\mathcal{I}_{f}}{\mathcal{I}_{i}}=\frac{\left\langle V\right\rangle }{N\ln\Delta}\label{eq:emergence}
\end{equation}
\[
\mathcal{S}=1-\mathcal{E}
\]
where $0\leq\left[\mathcal{E},\mathcal{S}\right]\leq1$. The minimum
energy of the system implies a potential energy which is equivalent
to the most self-organized system. $\mathcal{S}=1$ implies $\left\langle V\right\rangle =0$
and corresponds to a system where input and target are independent,
that as we discussed is an unexpected result. So, at the optimum,
for a bounded system, we will have $\mathcal{S}<1$.

López-Ruíz et al. \citep{lopez-ruiz_statistical_1995} defined complexity
as
\[
\mathcal{C}=\mathcal{S}\mathcal{E}
\]
From equations (\ref{eq:self-organization}) and (\ref{eq:emergence}),
we have
\begin{equation}
\mathcal{C}=\frac{\left\langle V\right\rangle }{N\ln\Delta}\left(1-\frac{\left\langle V\right\rangle }{N\ln\Delta}\right)\label{eq:Complexity}
\end{equation}
where $0\leq\mathcal{C}\leq1$. Equation (\ref{eq:Complexity}) allows
a comparison of the intrinsic complexity between different problems
based on the work done by the force field at the optimum, given the
scale invariance of $W$.

\subsection{Role of kinetic energy}

The MinMI principle provides a criterion that determines how the identification
of an optimal neural structure for a given problem can be found in
the minimum of mutual information between inputs and targets. However,
the equation (\ref{eq:Final state equation}) contains elements other
than the term for mutual information, result of having taken without
justification an eigenvalue equation having the same structure as
the Schrödinger equation. The obvious question is: why not just minimize
$I(\mathbf{t},\mathbf{x})$?

As empirical verification has been minimized the mutual information
containing a variational function given by equation (\ref{eq:yk})
and a set of normal probability densities, as described in Section
\ref{sec:The-potential}. This test showed that it's possible to found
a set of parameters $\Gamma$ that produce values for potential very
small, close to zero. However, networks obtained in this way do not
produce a significant correlation between the values of MI and the
error in the prediction of targets. The reason for this result has
already been commented in the previous sections: $I(t_{k},\mathbf{x})=0$
implies $p(t_{k},\mathbf{x})=p(t_{k})p(\mathbf{x})$, independence
between inputs and targets, and then the impossibility to build a
predictive model. It is necessary additional information that allows
to identify the minimum of mutual information which constitutes a
valid relation between data and represents an approximation to the
true relationship between inputs and targets. This additional information
is provided by the Laplacian in the kinetic energy term.

Perhaps the best way to understand the meaning of the Laplacian is
through a hydrodynamic analogy. Consider a function $\varphi$ as
the scalar potential of a irrotational compressible fluid.\footnote{Link between wave mechanics and hydrodynamics is well known since
the Madelung's derivations and related work, that show how Schrödinger\textquoteright s
equation in quantum mechanics can be converted into the Euler equations
for irrotational compressible flow \citep{chern_fluid_2017,madelung_quantum_1927,tsekov_bohmian_2012,tsekov_hydrodynamic_2019,vadasz_rendering_2016}.
However, the discussion in the text is simply a qualitative analogy
in order to understand the role of the Laplacian.} It is possible to define the velocity field of the fluid as the gradient
of the scalar potential, $\mathbf{v}=\nabla\varphi$, which is called
\emph{potential flow. }In this case, the Laplacian of the scalar potential
is nothing more than the divergence of the flow. For $\nabla^{2}\varphi\neq0$
in a certain point, then there exist an acceleration of the potential
field. In this sense the Laplacian can be seen as a \textquotedbl driving
force\textquotedbl .\footnote{This driving force has not to be confused with the force in classical
mechanics, which is given by the opposite of the gradient of a potential.} Furthermore, the Laplacian of a function at one point gives the difference
between the function value at that point and the average of the function
values in the infinitesimal neighborhood. Since the difference of
the average of surrounding and the point itself is actually related
to the curvature, the driving force can be considered as curvature
induced force. In this way, $\nabla^{2}\Psi$ is the divergence of
a gradient in the space of inputs and then may be associated with
the divergence of a flow given by the gradient of the probability
amplitude, which can be considered a diffusive term that conditioning
the concentration of the conditional measurements $x_{ij}$ \citep{stepanov_laplacian_2001}
and ultimately the probability $\left|\Psi\right|^{2}$. In this sense,
the kinetic term of the equation \ref{eq:Final state equation}) expresses
a constraint to the potential energy, which must be minimized compatibly
with a distribution of the conditional probability of the measurements
in the space $\mathbf{x}$.

To understand the nature of this constraint we analyze the mathematical
form of the kinetic term. By integrating the equation (\ref{eq:<T>})
over $\mathbf{x}$ we have
\begin{equation}
\begin{array}{rcl}
\left\langle T\right\rangle  & = & -\frac{2}{(2\pi)^{N/2}\left|\Sigma\right|^{1/2}}\sum_{i=1}^{N}\sigma_{i}\sum_{m=1}^{D}\sum_{n=1}^{D}c_{m}c_{n}\frac{\lambda_{m}\lambda_{n}}{\lambda_{m}+\lambda_{n}}\left(\frac{\pi}{\lambda_{m}+\lambda_{n}}\right)^{\frac{N}{2}}\times\\
 &  & \left[2\frac{\lambda_{m}\lambda_{n}}{\lambda_{m}+\lambda_{n}}(\eta_{im}-\eta_{in})^{2}-1\right]\prod_{i=1}^{N}\exp\left\{ -\frac{\lambda_{m}\lambda_{n}}{\lambda_{m}+\lambda_{n}}(\eta_{im}-\eta_{in})^{2}\right\} 
\end{array}\label{eq:<T> integrated}
\end{equation}
The sign of every term of the double sum on $(m,n)$ is given by the
product $c_{m}c_{n}\left[2\frac{\lambda_{m}\lambda_{n}}{\lambda_{m}+\lambda_{n}}(\eta_{im}-\eta_{in})^{2}-1\right]$.
Since experimentally it is found that the expected value of the kinetic
energy is positive for all the datasets studied and since $-\frac{2}{(2\pi)^{N/2}\left|\Sigma\right|^{1/2}}<0$,
there is a net effect given by the terms for which this product is
negative which leads to the condition
\[
-\frac{2}{(2\pi)^{N/2}\left|\Sigma\right|^{1/2}}\left(-c_{m}c_{n}\left[2\frac{\lambda_{m}\lambda_{n}}{\lambda_{m}+\lambda_{n}}(\eta_{im}-\eta_{in})^{2}-1\right]\right)>0
\]
and

\begin{equation}
\frac{\lambda_{m}\lambda_{n}}{\lambda_{m}+\lambda_{n}}(\eta_{im}-\eta_{in})^{2}>\frac{1}{2}\label{eq:T condition}
\end{equation}

We study the variation of the kinetic energy as a function of the
behavior of the following two factors present in the equation ((\ref{eq:<T> integrated}),
where we have separated the contribution of the $i$-th feature from
those for which $j\neq i$

\begin{equation}
f_{1}=\prod_{j\neq i}^{N}\exp\left\{ -\frac{\lambda_{m}\lambda_{n}}{\lambda_{m}+\lambda_{n}}(\eta_{jm}-\eta_{jn})^{2}\right\} \label{eq:f1}
\end{equation}
\begin{equation}
f_{2}=\left[2\frac{\lambda_{m}\lambda_{n}}{\lambda_{m}+\lambda_{n}}(\eta_{im}-\eta_{in})^{2}-1\right]\exp\left\{ -\frac{\lambda_{m}\lambda_{n}}{\lambda_{m}+\lambda_{n}}(\eta_{im}-\eta_{in})^{2}\right\} \label{eq:f2}
\end{equation}
In relation to $f_{1}$, $\left\langle T\right\rangle $ decreases
with the increase of $\frac{\lambda_{m}\lambda_{n}}{\lambda_{m}+\lambda_{n}}$
and distance $(\eta_{jm}-\eta_{jn})^{2}$. If we consider the $\lambda$
parameter as a measure of the variance $\sigma_{\lambda}^{2}=\frac{1}{2\lambda}$
associated with the correspondent Gaussian, we have $\frac{\lambda_{m}\lambda_{n}}{\lambda_{m}+\lambda_{n}}=\frac{1}{2(\sigma_{m}^{2}+\sigma_{n}^{2})}$
and the equality
\[
\exp\left\{ -\frac{\lambda_{m}\lambda_{n}}{\lambda_{m}+\lambda_{n}}(\eta_{jm}-\eta_{jn})^{2}\right\} =\exp\left\{ 2\frac{\sigma_{m}^{2}+\sigma_{n}^{2}}{(\eta_{jm}-\eta_{jn})^{2}}\right\} 
\]
The minimization of $\left\langle T\right\rangle $ therefore leads
to a decrease in dispersion and an increase in the centroids distance,
or in other words to the localization and separation of the Gaussians.
We can consider this behavior an equivalent of the Davies-Bouldin
index, which expresses the optimal balance between dispersion and
separation in clustering algorithms \citep{wajih_logarithmic_2017,xi_feature-level_2018}.
The issue has also been extensively studied in the context of RBF
networks \citep{benoudjit_width_2002,benoudjit_kernel_2003,majdisova_radial_2017,majdisova_radial_2018,wu_using_2012}.

\begin{figure}
\caption{\label{fig:f2}$f_{2}=\left[2\frac{\lambda_{m}\lambda_{n}}{\lambda_{m}+\lambda_{n}}(\eta_{im}-\eta_{in})^{2}-1\right]\exp\left\{ -\frac{\lambda_{m}\lambda_{n}}{\lambda_{m}+\lambda_{n}}(\eta_{im}-\eta_{in})^{2}\right\} $}

\begin{centering}
\includegraphics[width=1\textwidth]{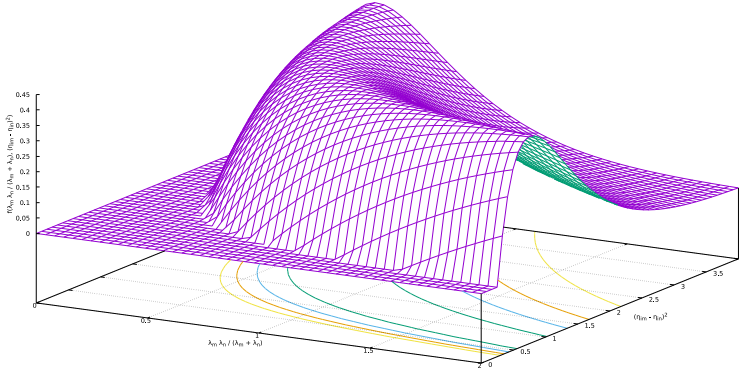}
\par\end{centering}
\end{figure}

For $f_{2}$ the behavior is more complex. Figure \ref{fig:f2} contains
the graphic representation of the surface given by $f_{2}$ with the
points that satisfy the condition (\ref{eq:T condition}). The domains
used for independent variables reflect the limits used in the tests:
\begin{itemize}
\item $\lambda\in[0:4]$, so $\frac{\lambda_{m}\lambda_{n}}{\lambda_{m}+\lambda_{n}}\in[0:2]$;
\item $\eta\in[-1:1]$ (the limits of the normalization range $\Delta$),
so $(\eta_{im}-\eta_{in})^{2}\in[0:4]$.
\end{itemize}
The qualitative trend of the kinetic energy can be summarized as follows:
\begin{enumerate}
\item \label{enu:f2-1}for large values of $\frac{\lambda_{m}\lambda_{n}}{\lambda_{m}+\lambda_{n}}$
(high values for $\frac{1}{2(\sigma_{m}^{2}+\sigma_{n}^{2})}$), $\left\langle T\right\rangle $
mainly decreases proportionally to the distance $(\eta_{im}-\eta_{in})^{2}$.
We have a behavior similar to that discussed for $f_{1}$;
\item \label{enu:f2-2}for small values of $\frac{\lambda_{m}\lambda_{n}}{\lambda_{m}+\lambda_{n}}$
(low values for $\frac{1}{2(\sigma_{m}^{2}+\sigma_{n}^{2})}$), $\left\langle T\right\rangle $
decreases inversely proportional to the distance $(\eta_{im}-\eta_{in})^{2}$.
\end{enumerate}
For high values of $\frac{\lambda_{m}\lambda_{n}}{\lambda_{m}+\lambda_{n}}$
there are very localized Gaussians whose centers must be as far apart
as possible so as to obtain a good representation of the entire domain
of the dataset. For low values of $\frac{\lambda_{m}\lambda_{n}}{\lambda_{m}+\lambda_{n}}$,
however, the Gaussians are very wide and their centers must be relatively
close so as not to lose part of their descriptive capacity, which
happens, for example, if the centroids are at the limits of the normalization
range. The predominant effect of the product $f_{1}f_{2}$ implies
a decrease of $\left\langle T\right\rangle $ with the increase of
the localization and the separation of the Gaussians that compose
the basis functions $\psi_{d}$ of the state function, with a small
modulation for low values of $\frac{\lambda_{m}\lambda_{n}}{\lambda_{m}+\lambda_{n}}$
as previously discussed. Note that this latter effect becomes less
important as the dimension of inputs, $N$, grows.

The role of kinetic energy is therefore to find the optimal balance
between variance and distribution of the Gaussian centers that make
up each basis function $\psi_{d}$ in order to obtain a good representation
of the domain defined by the dataset, and expresses a constraint for
potential energy. The minimization of the system energy obtained by
using a variational trial function in the state equation expresses
this balance. The optimal neural network obtained in this way represents
the structure with the minimum mutual information that contains an
adequate representation of the domain, and the kinetic operator $\hat{T}$
represents an additional term that prevents mutual information from
becoming too small and avoid to obtain solutions that do not contain
a relation between data.

\subsection{Operators}

All the quantities of interest in the model described can be calculated,
as happens in quantum mechanics, through the application of suitable
operators. However, it must be taken into consideration that the values
of the set of variational parameters $\Gamma$ obtained in the minimization
of energy are not, in general, valid for other observables when using
trial functions that are not the true eigenfunction of the system.
This does not invalidate the results obtained in this section, in
which we propose general expressions whose correctness in the numerical
results is subject to the use of the correct values of the parameters
$\Gamma$ for each observable under study.

\subsubsection{Expected value for output $y_{k}$}

Considering equation (\ref{eq:yk}) and (\ref{eq:Psi linear combination})
we can write the following expression for the expected value of $y_{k}$
\[
\left\langle y_{k}\right\rangle =\int y_{k}\left|\Psi\right|^{2}\,d\mathbf{x}=\sum_{p=1}^{P}\sum_{d=1}^{D}\sum_{l=1}^{D}w_{kp}c_{d}c_{l}\int\phi_{p}\psi_{d}\psi_{l}\,d\mathbf{x}+w_{k0}
\]
Using equations (\ref{eq:RBF function}) and (\ref{eq:psi(d) base function})
we obtain the final result
\[
\begin{array}{rcl}
\left\langle y_{k}\right\rangle  & = & w_{k0}+\sum_{p=1}^{P}\sum_{d=1}^{D}\sum_{l=1}^{D}w_{kp}c_{d}c_{l}\left(\frac{\pi}{\xi_{p}+{\it \lambda}_{l}+{\it \lambda}_{d}}\right)^{\frac{N}{2}}\times\\
 &  & \prod_{i=1}^{N}\exp\left\{ -\frac{\xi_{p}\left[\lambda_{l}\left({\it \omega}_{pi}-\eta_{il}\right)^{2}+{\it \lambda}_{d}\left({\it \omega}_{pi}-\eta_{id}\right)^{2}\right]+\lambda_{d}\lambda_{l}\left(\eta_{il}-\eta_{id}\right)^{2}}{\xi_{p}+{\it \lambda}_{l}+{\it \lambda}_{d}}\right\} 
\end{array}
\]

\subsubsection{Expected value for $x_{i}$}

The expected value of a component $x_{i}$ of of the conditional probability
$\left|\Psi(\mathbf{x})\right|^{2}$ is given by
\[
\left\langle x_{i}\right\rangle =\int x_{i}\left|\Psi\right|^{2}\,d\mathbf{x}=\sum_{d=1}^{D}\sum_{l=1}^{D}c_{d}c_{l}\int x_{i}\psi_{d}\psi_{l}\,d\mathbf{x}
\]
and using equations (\ref{eq:RBF function}) and (\ref{eq:psi(d) base function})
\[
\left\langle x_{i}\right\rangle =\sum_{d=1}^{D}\sum_{l=1}^{D}c_{d}c_{l}\left(\lambda_{d}\eta_{id}+\lambda_{l}\eta_{il}\right)\frac{\pi^{\frac{N}{2}}}{\left(\lambda_{d}+\lambda_{l}\right)^{\frac{N+2}{2}}}\prod_{i=1}^{N}\exp\left\{ -\frac{\lambda_{d}\lambda_{l}}{\lambda_{d}+\lambda_{l}}\left(\eta_{il}-\eta_{id}\right)^{2}\right\} 
\]

\subsubsection{Expected value for the variance of $x_{i}$}

Variance for $x_{i}$ is given by
\[
(\Delta x_{i})^{2}=\left\langle x_{i}^{2}\right\rangle -\left\langle x_{i}\right\rangle ^{2}
\]

\subsection{Uncertainty principle}

In this section we make an analysis of the possible validity of the
uncertainty principle within the EANNs. For simplicity we will consider
in the treatment the one-dimensional case, that is a system composed
by a neural network with a single input $x$.

From a direct comparison between the differential equation describing
the physical behavior of the EANNs and the Schrõdinger equation, we
postulate the following expression for the operator $\hat{p}_{x}^{2}$

\begin{equation}
\hat{p}_{x}^{2}=-\sigma_{x}^{2}\nabla^{2}\label{eq:Operatore px^2}
\end{equation}
The negative sign allows to obtain positive kinetic energies, as has
been confirmed by the experiments detailed in Section \ref{sec:Results},
and necessarily leads to an expression for momentum operator $\hat{p}_{x}$
which contains the imaginary unit
\begin{equation}
\hat{p}_{x}=\frac{\sigma_{x}}{i}\nabla\label{eq:Operatore px}
\end{equation}

According to the laws of probability, the variances $\left(\Delta x\right)^{2}$
and $\left(\Delta p_{x}\right)^{2}$ related to the observables $x$
and $p_{x}$ can be written as \citep{levine_quantum_2014}
\begin{equation}
\left(\Delta x\right)^{2}=\left\langle (x-\left\langle x\right\rangle )^{2}\right\rangle =\left\langle x^{2}\right\rangle -\left\langle x\right\rangle ^{2}=\int\Psi^{*}(\hat{x}-\left\langle x\right\rangle )^{2}\Psi\,dx\label{eq:var(x)}
\end{equation}
\begin{equation}
\left(\Delta p_{x}\right)^{2}=\left\langle (p_{x}-\left\langle p_{x}\right\rangle )^{2}\right\rangle =\left\langle p_{x}^{2}\right\rangle -\left\langle p_{x}\right\rangle ^{2}=\int\Psi^{*}(\hat{p}_{x}-\left\langle p_{x}\right\rangle )^{2}\Psi\,dx\label{eq:var(px)}
\end{equation}
Equations (\ref{eq:var(x)}) and (\ref{eq:var(px)}) are in fact definitions
of a variance. In particular, the equation (\ref{eq:var(x)}) is the
expected value of the quadratic deviation of the variable $x$ measured
against a conditional probability density given by the square of the
state function, and should not be confused with $\sigma_{x}^{2}$,
which is the variance given by the marginal probability density of
$x$. Taking into account the definition of expected value and the
mathematical properties of Schwartz's inequality, for the product
of standard deviations we have \citep{briggs_derivation_2008,robertson_uncertainty_1929}
\begin{equation}
\Delta x\Delta p_{x}\geq\frac{1}{2}\left|\int\Psi^{*}[\hat{x},\hat{p}_{x}]\Psi\,dx\right|\label{eq:Uncertainty principle}
\end{equation}
where $[\hat{x},\hat{p}_{x}]$ is the commutator for position, $\hat{x}$,
and momentum, $\hat{p}_{x}$, operators.

If we consider valid the rules of commutation in quantum mechanics
and their meaning as the possibility or not of measuring with arbitrary
precision the values of conjugated variables, then there exits an
uncertainty relationship also in the EANNs. The reason lies in the
fact that any operator composed of a first derivative does not commute
with the position operator. By defining the operator $\hat{D}=\frac{d}{dx}$,
the commutator with $\hat{x}$ gives rise, as is known, to
\[
\hat{D}\hat{x}=\hat{1}+\hat{x}\hat{D}
\]
\[
\left[\frac{d}{dx},\hat{x}\right]=\hat{D}\hat{x}-\hat{x}\hat{D}=\hat{1}
\]
where $\hat{1}$ is the unit operator. In our case, for the position-momentum
commutator, taking into account that $\left[\hat{x},\hat{p}_{x}\right]=-\left[\hat{p}_{x},\hat{x}\right]$,
we have
\[
\left[\hat{x},\hat{p}_{x}\right]=\left[\hat{x},\frac{\sigma_{x}}{i}\frac{\partial}{\partial x}\right]=\frac{\sigma_{x}}{i}\left[\hat{x},\frac{\partial}{\partial x}\right]=-\frac{\sigma_{x}}{i}\left[\frac{\partial}{\partial x},\hat{x}\right]=-\frac{\sigma_{x}}{i}=i\sigma_{x}
\]
Finally, from the equation (\ref{eq:Uncertainty principle}) 
\[
\Delta x\Delta p_{x}\geq\frac{1}{2}\left|\int\Psi^{*}i\sigma_{x}\Psi\,dx\right|=\frac{\sigma_{x}}{2}\left|i\right|\left|\int\Psi^{*}\Psi\,dx\right|
\]
\begin{equation}
\Delta x\Delta p_{x}\geq\frac{\sigma_{x}}{2}\label{eq:Principio indeterminazione final}
\end{equation}
Assuming therefore an operator of the form (\ref{eq:Operatore px})
and taking into account very general considerations, we obtain an
equivalent of the uncertainty principle for the EANNs. This result
is not surprising given the mathematical equivalence with the Schrõdinger
equation.

It is possible to generalize this result for pairs of generic conjugated
variables if one assumes the validity for the EANN of some classical
results in quantum mechanics, which however needs formal verification.
If the equation (\ref{eq:Final state equation}) governs the behavior
of a true quantum system in which not all pairs of variables commute,
we can hypothesize the existence of a classic system in which all
variables instead commute. In this context, from a quantum point of
view an equivalent of Dirac's proposal for pairs of variables $(f,g)$
is
\[
[\hat{f},\hat{g}]=i\sigma_{x}\left\{ f,g\right\} 
\]
where $\left\{ f,g\right\} $ is the Poisson bracket between $f$
and $g$.

From a physical point of view, the equation (\ref{eq:Uncertainty principle})
is a formalization of the impossibility of simultaneously measuring
position and momentum with arbitrary precision, ie the fact that the
variance in the position is subject to a kinematic constraint \citep{bastos_robertson-schrodinger_2015}.
Equivalently, within non-commutative algebras, it makes explicit the
fact that position and momentum cannot be made independent from a
statistical point of view, constituting a limit to a priori knowledge
regarding observable statistics and their predictability. Equation
(\ref{eq:Principio indeterminazione final}) highlights how the product
in the uncertainties or variances of two non-independent observables
cannot be less than a certain minimum value, which is related to the
standard deviation of the marginal density of the dataset.

Equation (\ref{eq:Principio indeterminazione final}) can be derived
from purely statistical considerations. Starting from the inequality
proposed independently by Bourret \citep{bourret_note_1958}, Everett
\citep{dewitt_many-worlds_1973}, Hirschman \citep{hirschman_note_1957}
and Leipnick \citep{leipnik_entropy_1959}, satisfied by each function
$\psi$ and its Fourier transform $\widetilde{\psi}$

\[
-\int\left|\psi(x)\right|^{2}\ln\left|\psi(x)\right|^{2}\,dx-\int\left|\widetilde{\psi}(p_{x})\right|^{2}\ln\left|\widetilde{\psi}(p_{x})\right|^{2}\,dp_{x}=h(x)+h(p_{x})\geq1+\ln\pi
\]
Beckner \citep{beckner_inequalities_1975} and Bialinicki-Birula and
Micielski \citep{bialynicki-birula_uncertainty_1975} demonstrated
\begin{equation}
h(x)+h(p_{x})\geq\ln(\pi e\sigma_{x})\label{eq:principio di indeterminazione come entropia differenziale}
\end{equation}
where we assumed the equivalence $\hbar\equiv\sigma_{x}$. Since the
differential entropy of the normal probability density is maximum
among all distributions with the same variance, for a generic probability
density $f(z)$ we can write the inequality
\begin{equation}
h(z)\leq\ln(\sqrt{2\pi e}\Delta z)\label{eq:entropia differenziale densit=0000E0 normale}
\end{equation}
Substituting (\ref{eq:entropia differenziale densit=0000E0 normale})
in (\ref{eq:principio di indeterminazione come entropia differenziale})
we have
\[
\Delta x\Delta p\geq\frac{1}{2\pi e}\exp\left\{ h(x)+h(p_{x})\right\} \geq\frac{\sigma_{x}}{2}
\]
It is worth mentioning that the equation (\ref{eq:principio di indeterminazione come entropia differenziale})
is derived from mathematical properties and acquires the form of the
text at the end of the proof, when considering a concrete expression
for $p_{x}$.

Finally, the uncertainty principle can also be derived from the Cramér-Rao
bound, as demonstrated by several authors \citep{angelow_evolution_2009,dembo_information_1991,frowis_tighter_2015,gibilisco_robertson-type_2008,gibilisco_uncertainty_2007,hall_prior_2004,parthasarathy_philosophy_2009,rodriguez_disturbance-disturbance_2018,theodoridis_machine_2015},
that is, starting from exclusively statistical properties. Section
\ref{sec:Quantum-mechanics-and information theory} contains the bibliographic
references of some significant works related to this topic.

\subsection{Time-dependent system}

The discussion of previous sections has considered stationary systems.
Independence from time, as we have highlighted, is based on the temporal
invariance of the set of constants $\vec{\rho},\vec{\theta},\vec{\mu}$
and $\vec{\sigma}$ which identify the problem. However, given a dataset,
it is reasonable to assume its temporal evolution, identified as the
acquisition of new data that vary $\vec{\rho},\vec{\theta},\vec{\mu}$
and $\vec{\sigma}$ between an initial state and a final state. In
the following we will consider for simplicity the one-dimensional
case ($N=1$).

We postulate the following expression for a time-dependent system\footnote{Also in quantum mechanics the wave equation is postulated and not
demonstrated starting from considerations on the classic wave equation.}\nomenclature[tau]{$\tau$}{time}\nomenclature[tau']{$\tau '$}{logarithmic time}
\begin{equation}
\begin{array}{rcl}
-\frac{\mathcal{A}}{i}\frac{d\Psi(x,\tau)}{d\tau} & = & -\frac{\sigma_{x}^{2}(x,\tau)}{(2\pi)^{1/2}\left|\Sigma(x,\tau)\right|^{1/2}}\nabla^{2}\Psi(x,\tau)+\mathcal{N}(x,\tau)\times\\
 &  & \sum_{k=1}^{C}\left[\alpha_{k}(x,\tau)y_{k}^{2}(x,\tau)+\beta_{k}(x,\tau)y_{k}(x,\tau)+\gamma_{k}(x,\tau)\right]\Psi(x,\tau)
\end{array}\label{eq:time-dependent state function}
\end{equation}
where $\tau$ is time, $\mathcal{A}$ is a factor to be determined
and where we have explicitly highlighted the temporal dependence of
all terms. In this version, the values of $\vec{\rho},\vec{\theta},\vec{\mu}$
and $\vec{\sigma}$ and the weights of the network are generally different
in the initial and final states and therefore depend on time in a
way, however, that we don't know. This also implies a temporal evolution
in the marginal densities $p(x)$ and $p(t)$. To make matters worse
than quantum physical systems that have time-dependent potentials,
the equation (\ref{eq:time-dependent state function}) also contains
a time-dependent factor for the kinetic term ($\sigma,\Sigma)$. Furthermore,
in the absence of explicit expressions, the time functional factors
introduce dimensional problems in the equation (\ref{eq:time-dependent state function}).
In practice, given the lack of knowledge of the functional dependence
on time of the terms that appear on the right hand side of the equation
(\ref{eq:time-dependent state function}), we can consider it as not
solvable. Imaginary unit in the left hand side is necessary if we
consider certain conditions met, as we will see later in the discussion.
Term $-\frac{\mathcal{A}}{i}\frac{d\Psi}{d\tau}$ can be seen as the
result of a Wick rotation of $\mathcal{A}\frac{d\Psi}{d\tau_{w}}$
in time $\tau_{w}=-i\tau$.

Systems with time-dependent potentials are approached in quantum mechanics,
in many cases, with a Hamiltonian consisting of two terms: one time-independent
and one time-dependent, the last being treated as a perturbation
\[
H(x,\tau)=H_{0}(x)+V(x,\tau)
\]
where $H_{0}(x)$ is the Hamiltonian of the equation (\ref{eq:Final state equation}).
Supposing that $V(x,\tau)$ is negligible compared to $H_{0}(x)$,\footnote{This assumption lies in the value of the ratio between kinetic and
potential energy at the optimum in the version time-independent of
the formalism, where it occurs $\frac{\left\langle T\right\rangle }{\left\langle V\right\rangle }\gg1$.} equation (\ref{eq:time-dependent state function}) becomes
\begin{equation}
-\frac{\mathcal{A}}{i}\frac{d\Psi}{d\tau}=-\frac{\sigma_{x}^{2}}{(2\pi)^{1/2}\left|\Sigma\right|^{1/2}}\nabla^{2}\Psi+\mathcal{N}(\mu,\sigma_{x}^{2})\sum_{k=1}^{C}(\alpha_{k}y_{k}^{2}+\beta_{k}y_{k}+\gamma_{k})\Psi\label{eq:time-dependent state function separable}
\end{equation}
The right hand side has units given by the factor $\frac{1}{\left|\Sigma\right|^{1/2}}$
and must be introduced in the left hand side to maintain dimensional
consistency. However, the differential equation governing an EANN
does not contain any time-dependent constant factor, unlike what happens
with the constant $\hbar$ in the time-dependent version of the Schrõdinger
equation. The dimensional coherence between the two members of the
equation (\ref{eq:time-dependent state function separable}) imposes
the following expression for $\mathcal{A}$
\[
\mathcal{A}=\frac{\tau}{\left|\Sigma\right|^{1/2}}
\]

Assuming a solution given in the form
\[
\Psi(x,\tau)=f(\tau)\psi(x)
\]
equation (\ref{eq:time-dependent state function separable}) becomes
separable, being the parts time-dependent and time-independent equal
to a constant that can be shown is the system energy. For the time-dependent
part we have the following ordinary differential equation
\[
\frac{df(\tau)}{f(\tau)}=-\frac{i\left|\Sigma\right|^{1/2}E}{\tau}\,d\tau
\]
whose solution is
\begin{equation}
f(\tau)=\tau^{-i\left|\Sigma\right|^{1/2}E}=\exp\left\{ -i\left|\Sigma\right|^{1/2}E\ln(\tau)\right\} \label{eq:f(tau)}
\end{equation}
where the integration constant has been omitted since it can be included
as a factor in the $\psi(x)$ function. The last equality of equation
(\ref{eq:f(tau)}) expresses a periodic function in the variable $\tau'=\ln(\tau)$.

We now justify the presence of the imaginary unit in the equation
(\ref{eq:f(tau)}). Following quantum mechanics, we impose the condition
that for a pure eigenfunction $\Psi(x,\tau)$ the probability density
is stationary \citep{levine_quantum_2014}
\[
\left|\Psi(x,\tau)\right|^{2}=\left|\psi(x)\right|^{2}
\]
Mathematically, this condition leads to
\begin{equation}
f^{*}(\tau)f(\tau)=1\label{eq:f*(tau)f(tau)=00003D1}
\end{equation}
which can only be satisfied if $f(\tau)$ is a complex function. More
formally, it is possible to define an operator $\hat{U}(\tau,\tau_{0})$
that describes the temporal evolution of the system according to the
equation
\[
-\frac{\tau}{i\left|\Sigma\right|^{1/2}}\frac{d}{d\tau}\hat{U}(\tau,\tau_{0})=\hat{H}\hat{U}(\tau,\tau_{0})
\]
subject to the initial condition
\[
\hat{U}(\tau_{0},\tau_{0})=1
\]
Equality (\ref{eq:f*(tau)f(tau)=00003D1}) expresses the unitarity
of $\hat{U}$, a property derived from the hermiticity of $\hat{H}$
\citep{messiah_quantum_1966}.

As is known, the probability density is not stationary for a system
that makes a temporal transition from a state $m$ to a state $n$.
The state function of this system can be written as a linear combination
of the two states
\[
\text{\ensuremath{\Psi_{mn}(x,\tau)=c_{m}\Psi_{m}(x,\tau)+c_{n}\Psi_{n}(x,\tau)}}
\]
If $(c_{m},c_{n})\neq0$, the probability density is
\[
\left|\Psi_{mn}\right|^{2}=c_{m}^{2}\left|\Psi_{m}\right|^{2}+c_{n}^{2}\left|\Psi_{n}\right|^{2}+c_{m}c_{n}(\Psi_{m}^{*}\Psi_{n}+\Psi_{m}\Psi_{n}^{*})
\]
The first two terms of the right hand side are time-independent. By
developing the calculation and considering, as we did in this paper,
only real functions for the time-independent component of the state
function, the final time-dependent probability density is
\begin{equation}
\left|\Psi_{mn}\right|^{2}=c_{m}^{2}\psi_{m}^{2}+c_{n}^{2}\psi_{m}^{2}+2c_{m}c_{n}\psi_{m}\psi_{n}\cos\left\{ \left(\left|\Sigma_{m}\right|^{1/2}E_{m}-\left|\Sigma_{n}\right|^{1/2}E_{n}\right)\tau'\right\} \label{eq:Psi(x,tau)^2}
\end{equation}
Equation (\ref{eq:Psi(x,tau)^2}) expresses the evolution of the conditional
probability density $\Psi(x,\tau|\mathbf{t})$ between two states
separated in time. The state function consisting of a mixture of two
energy states does lead to a conditional density that oscillates in
the logarithmic time $\tau'$ with frequency\nomenclature[nu]{$\nu$}{frequency}
\[
\nu=\left|\Sigma_{m}\right|^{1/2}E_{m}-\left|\Sigma_{n}\right|^{1/2}E_{n}
\]

\section{\label{sec:Results}Results}

The resolution of the system (\ref{eq:Secular system}) requires considerable
computational powers. For this reason the minimum energy was calculated
with a genetic algorithm (GA).

The test problem comes from the Statlib repository.\footnote{\url{http: //lib.stat.cmu.edu/datasets/}}
It is a synthetic dataset made up of 3848 records, generated by David
Coleman, referred to for convenience as POLLEN, which represents geometric
and physical characteristics of pollen grain samples. It consists
of 5 variables: the first three are the lengths in the directions
x (ridge), y (nub) and z (crack), the fourth is the weight and the
fifth is the density, the latter being the target of the problem.
In our model they represent, respectively, $x_{1}$, $x_{2}$, $x_{3}$,
$x_{4}$ and $t_{1}$. The choice of this problem lies in the fact
that the data were generated with normal distributions with low correlations,
and is therefore close to the initial assumptions of the model for
$\mathbf{x}$ and $\mathbf{t}$. Tables \ref{tab:POLLEN general}
and \ref{tab:POLLEN correlation matrix} show the general statistics
of the dataset.

\begin{table*}
\caption{\label{tab:POLLEN general}POLLEN dataset, general features. The table
shows the means ($\mu$, $\rho$), standard deviations ($\sigma$,
$\theta$), skewness and kurtosis (the reference for normality is
0) of original and normalized data}

\centering{}%
\begin{tabular}{c|rr|rr|rr}
\hline 
 & \multicolumn{2}{c|}{Original data} & \multicolumn{2}{r|}{Normalized data} & \multirow{2}{*}{Skewness} & \multirow{2}{*}{Kurtosis}\tabularnewline
\cline{1-5} \cline{2-5} \cline{3-5} \cline{4-5} \cline{5-5} 
Var & $\mu$ / $\rho$ & $\sigma$ / $\theta$ & $\mu$ / $\rho$ & $\sigma$ / $\theta$ &  & \tabularnewline
\hline 
\hline 
$x_{1}$ & -3.637e-03 & 6.398 & 0.0418 & 0.2863 & -0.130 & -0.057\tabularnewline
$x_{2}$ & 1.597e-04 & 5.186 & -0.0257 & 0.3082 & 0.072 & -0.311\tabularnewline
$x_{3}$ & 3.103e-03 & 7.875 & 0.0178 & 0.2551 & -0.057 & -0.158\tabularnewline
$x_{4}$ & 4.237e-03 & 10.004 & -0.0252 & 0.2876 & 0.109 & -0.163\tabularnewline
$t_{1}$ & 1.662e-04 & 3.144 & 0.0512 & 0.2745 & 0.110 & 0.192\tabularnewline
\hline 
\end{tabular}
\end{table*}

\begin{table}
\caption{\label{tab:POLLEN correlation matrix}Dataset POLLEN, correlation
matrix}

\centering{}%
\begin{tabular}{c|rrrrr}
\hline 
 & $x_{1}$ & $x_{2}$ & $x_{3}$ & $x_{4}$ & $t_{1}$\tabularnewline
\hline 
\hline 
$x_{1}$ & 1.00 & 0.13 & -0.13 & -0.90 & -0.57\tabularnewline
$x_{2}$ & 0.13 & 1.00 & 0.08 & -0.17 & 0.33\tabularnewline
$x_{3}$ & -0.13 & 0.08 & 1.00 & 0.27 & -0.15\tabularnewline
$x_{4}$ & -0.90 & -0.17 & 0.27 & 1.00 & 0.24\tabularnewline
$t_{1}$ & -0.57 & 0.33 & -0.15 & 0.24 & 1.00\tabularnewline
\hline 
\end{tabular}
\end{table}

The characteristics of the genetic algorithm have been described in
a previous paper \citep{FranciscoYepesBarrera-2007-385}. This is
a steady-state GA, with a generation gap of one or two, depending
on the operator applied. The population has binary coding and implements
a fitness sharing mechanism \citep{Gao:MTMBNGA:2006} to allow speciation
and avoid premature convergence, according to the equations
\begin{equation}
E_{l}^{'}=E_{l}\sum_{m}\varphi(d_{lm})\label{eq:niche sharing}
\end{equation}
\[
\varphi(d_{lm})=\left\{ \begin{array}{lcr}
1-\left(\frac{d_{lm}}{R}\right)^{\upsilon} & \Rightarrow & d_{kl}<R\\
0 & \Rightarrow & d_{kl}\geq R
\end{array}\right.
\]
being $\varphi(d_{lm})$ a function of the diversity between individuals
$l$ and $m$, $d_{lm}$ the Hamming distance and $R$ the niche radius
within which individuals are considered similar. Niche sharing implements
a correction to energy calculated based on the similarity between
the individual $l$ and the rest of the population. The more similar
it is, the greater the value of $\varphi(d_{lm})$, penalizing the
energy in the equation (\ref{eq:niche sharing}) since we are minimizing.

The decoding of the genotype implements the Gray code to avoid discontinuities
in the binary representation. The transformation between the binary
representations, $b$, and Gray, $g$, for the $i$-th bit, considering
numbers composed of $n$ bits numbered from right to left, with the
most significant bit on the left, is given by
\[
\begin{array}{c}
g_{i}=\left\{ \begin{array}{ccl}
b_{i} & \Rightarrow & i=n\\
b_{i+1}\otimes b_{i} & \Rightarrow & i<n
\end{array}\right.\\
b_{i}=\left\{ \begin{array}{ccl}
g_{i} & \Rightarrow & i=n\\
b_{i+1}\otimes g_{i} & \Rightarrow & i<n
\end{array}\right.
\end{array}
\]
where $\otimes$ is the XOR operator.

The GA uses four operators: crossover, mutation, uniform crossover
and internal crossover, and performs a search in the space of the
computed energies according to the equation (\ref{eq:Energy expected value}),
but simultaneously realizes a search in the space of the operators
through the use of two additional bits in the genotype of each individual
of the population. This allows a dynamic choice of the probabilities
of each operator at each moment of the calculation, according to the
fraction of elements of the population that were generated and encode
for each of the four possibilities. The initial population is randomly
generated.

The procedure for assessing an individual consists of the following
steps:
\begin{enumerate}
\item the values of the $n_{\Gamma}$ parameters are generated within certain
prefixed ranges through the application of one of the operators;
\item the network output, $y_{k}$, is generated for each element of the
dataset. This set of values allows to calculate $\chi_{k}$;
\item the $D\times D$ elements of the matrices $\mathbf{H}$ and $\mathbf{S}$
are computed by means of the integrals (\ref{eq:Hmn}) and (\ref{eq:Smn});
\item the determinant (\ref{eq:Secular determinant}) is calculated;
\item the system (\ref{eq:Secular system}) is solved.
\end{enumerate}
Result is the $D$ energy values, $E_{d}$, and the $D$ coefficients
$c$ for each of the $D$ state functions $\Psi_{d}$. The lower value
among $E_{d}$ represents the global optimum of the problem.

Before the execution of the tests, a preprocessing of the dataset
was performed, normalizing $\mathbf{x}$ and $t$ within the range
$\Delta\equiv[-1:1]$. 15 calculations were conducted, each consisting
of 10 concurrent processes sharing the best solution found. In each
calculation the set of lower energy solutions found in the previous
calculations were introduced. The values of the $n_{\Gamma}$ parameters
were varied within certain pre-established ranges, identified through
a preliminary test campaign. The reference ranges are shown in Table
\ref{tab:Test range}. Table \ref{tab:GA parameters} shows the reference
values of the parameters used in the genetic algorithm.

\begin{table}
\caption{\label{tab:Test range}Values of the model and range of variability
of the $n_{\Gamma}$ variational parameters, and $\mathbf{x}$ and
$t_{1}$ data of the dataset}

\centering{}%
\begin{tabular}{c|c}
\hline 
Variable & Value\tabularnewline
\hline 
\hline 
C & 1\tabularnewline
D & 12\tabularnewline
N & 4\tabularnewline
P & 20\tabularnewline
$\mathbf{x},t_{1}$ & $[-1:1]$\tabularnewline
$\lambda,\xi$ & $[0:4]$\tabularnewline
w & $[-4:4]$\tabularnewline
$\eta,\omega$ & $[-1:1]$\tabularnewline
\hline 
\end{tabular}
\end{table}

\begin{table}
\caption{\label{tab:GA parameters}Reference values of the genetic algorithm}

\centering{}%
\begin{tabular}{l|c}
\hline 
Variable & Value\tabularnewline
\hline 
\hline 
Population & 250\tabularnewline
Point mutation probability & $[0:0.01]$\tabularnewline
$\upsilon$ & 1\tabularnewline
Chromosome length & 3877 bits\tabularnewline
Calculation cycles & 20000\tabularnewline
\hline 
\end{tabular}
\end{table}

\begin{sidewaystable*}
\caption{\label{tab:Net parameters}Results of the genetic algorithm for the
parameters of basis functions $\phi$ of the network $y_{k}$}

\centering{}%
\begin{tabular}{c||c|cccc}
\hline 
$\xi$ & P \textbackslash{} N & $\omega_{1}$ & $\omega_{2}$ & $\omega_{3}$ & $\omega_{4}$\tabularnewline
\hline 
\hline 
2.919328e+00 & $\omega_{1}$ & -9.616850e-01 & 5.699350e-01 & -1.007076e-01 & 8.523940e-01\tabularnewline
8.423300e-01 & $\omega_{2}$ & 9.177860e-01 & 1.001960e-01 & 7.096600e-01 & 4.823590e-01\tabularnewline
2.087150e-01 & $\omega_{3}$ & -2.578000e-01 & 5.203320e-01 & -8.117540e-01 & -8.377960e-01\tabularnewline
3.914241e+00 & $\omega_{4}$ & -3.191230e-01 & 4.871910e-01 & -8.830230e-01 & 5.089810e-01\tabularnewline
5.823370e-01 & $\omega_{5}$ & -6.928290e-01 & 2.672930e-01 & -4.297810e-01 & -3.016050e-01\tabularnewline
1.926436e+00 & $\omega_{6}$ & -8.767050e-01 & -3.759150e-01 & -9.127510e-01 & 9.698310e-01\tabularnewline
3.097580e-01 & $\omega_{7}$ & -5.724300e-01 & 9.825380e-01 & -6.680690e-01 & 9.482790e-01\tabularnewline
3.858523e+00 & $\omega_{8}$ & 2.538800e-01 & 8.884770e-01 & 8.543280e-01 & 5.800880e-01\tabularnewline
2.705241e+00 & $\omega_{9}$ & -7.541110e-01 & -1.327950e-01 & -5.627170e-01 & 9.081580e-01\tabularnewline
9.479430e-01 & $\omega_{10}$ & -6.158020e-01 & 1.714950e-01 & -9.497280e-01 & 9.030120e-01\tabularnewline
1.103653e+00 & $\omega_{11}$ & -4.964920e-01 & 1.040151e-01 & -9.430600e-01 & 3.068200e-01\tabularnewline
3.384321e+00 & $\omega_{12}$ & 6.543940e-01 & -9.356290e-01 & 4.894600e-01 & 7.890730e-01\tabularnewline
7.454340e-01 & $\omega_{13}$ & -8.030510e-01 & -2.275000e-01 & -4.027400e-01 & -5.026880e-01\tabularnewline
1.926320e+00 & $\omega_{14}$ & -5.985470e-01 & 1.618350e-01 & -8.644710e-01 & -2.648060e-01\tabularnewline
1.349549e+00 & $\omega_{15}$ & -2.659290e-01 & -4.931090e-01 & 7.594300e-01 & 1.028353e-01\tabularnewline
2.424388e+00 & $\omega_{16}$ & 5.962460e-01 & 6.320610e-01 & -4.129300e-01 & -5.073280e-01\tabularnewline
6.307950e-01 & $\omega_{17}$ & -4.769380e-01 & 8.363110e-01 & 2.817150e-01 & 1.914970e-01\tabularnewline
2.245460e-01 & $\omega_{18}$ & 5.409380e-01 & 7.000500e-01 & 9.406210e-01 & -3.607610e-01\tabularnewline
3.103390e+00 & $\omega_{19}$ & -1.407570e-01 & -1.708480e-01 & -3.897940e-01 & 1.832320e-01\tabularnewline
2.451465e+00 & $\omega_{20}$ & -9.182390e-01 & 3.936930e-01 & 3.930220e-01 & 4.795600e-01\tabularnewline
\hline 
\end{tabular}
\end{sidewaystable*}

\begin{table}
\caption{\label{tab:Weights}Results of the genetic algorithm for network weights}

\centering{}%
\begin{tabular}{c|c}
\hline 
P \textbackslash{} C & $w_{1}$\tabularnewline
\hline 
\hline 
$w_{0}$ & 1.638795e+00\tabularnewline
$w_{1}$ & 1.419237e+00\tabularnewline
$w_{2}$ & -1.858620e+00\tabularnewline
$w_{3}$ & -2.603692e+00\tabularnewline
$w_{4}$ & 1.118712e+00\tabularnewline
$w_{5}$ & 1.239672e+00\tabularnewline
$w_{6}$ & -1.197724e+00\tabularnewline
$w_{7}$ & -1.490210e+00\tabularnewline
$w_{8}$ & 1.755702e+00\tabularnewline
$w_{9}$ & 9.814800e-01\tabularnewline
$w_{10}$ & -2.964685e+00\tabularnewline
$w_{11}$ & 2.344582e+00\tabularnewline
$w_{12}$ & -1.652975e+00\tabularnewline
$w_{13}$ & 3.030500e-01\tabularnewline
$w_{14}$ & -6.988870e-01\tabularnewline
$w_{15}$ & -3.270600e-01\tabularnewline
$w_{16}$ & 5.944650e-01\tabularnewline
$w_{17}$ & 1.834500e+00\tabularnewline
$w_{18}$ & -5.652970e-01\tabularnewline
$w_{19}$ & -1.576270e-01\tabularnewline
$w_{20}$ & -9.845600e-01\tabularnewline
\hline 
\end{tabular}
\end{table}

\begin{sidewaystable*}
\caption{\label{tab:Psi parameters}Results of the genetic algorithm for the
coefficients and parameters of basis functions $\psi_{d}$ of the
state function $\Psi$}

\centering{}%
\begin{tabular}{cc||c|cccc}
\hline 
$c$ & $\lambda$ & P \textbackslash{} N & $\eta_{1}$ & $\eta_{2}$ & $\eta_{3}$ & $\eta_{4}$\tabularnewline
\hline 
\hline 
1.553121e+00 & 1.390160e-01 & $\eta_{1}$ & -1.231300e-01 & -4.406100e-01 & -2.818570e-01 & 2.351560e-01\tabularnewline
-2.359102e-01 & 1.925760e-01 & $\eta_{2}$ & -1.194100e-01 & 1.004575e-01 & -1.022693e-01 & 1.308700e-01\tabularnewline
-1.904793e+00 & 1.085620e-01 & $\eta_{3}$ & 8.793090e-01 & 4.690850e-01 & -3.208850e-01 & 1.048543e-01\tabularnewline
-2.639326e+00 & 1.093980e-01 & $\eta_{4}$ & -4.300890e-01 & -4.877940e-01 & -1.046653e-01 & 2.051050e-01\tabularnewline
2.014761e-01 & 1.000033e-01 & $\eta_{5}$ & -8.999490e-01 & 1.039123e-01 & -4.522900e-01 & 4.788520e-01\tabularnewline
1.682093e-01 & 1.394260e-01 & $\eta_{6}$ & -7.989670e-01 & 1.157740e-01 & 4.029300e-01 & -1.007880e-01\tabularnewline
-2.765379e-01 & 1.799000e-01 & $\eta_{7}$ & 1.007745e-01 & -6.873660e-01 & -4.912310e-01 & 2.924720e-01\tabularnewline
9.654897e-01 & 1.000023e-01 & $\eta_{8}$ & -5.837550e-01 & -7.069950e-01 & 1.034500e-01 & 3.353310e-01\tabularnewline
4.245192e-01 & 1.382240e-01 & $\eta_{9}$ & 9.013570e-01 & 6.677700e-01 & -2.735550e-01 & 1.025390e-01\tabularnewline
8.729764e-01 & 1.000193e-01 & $\eta_{10}$ & 9.922380e-01 & 4.842470e-01 & -4.106250e-01 & 2.306550e-01\tabularnewline
1.726046e-01 & 1.000000e-01 & $\eta_{11}$ & -7.864230e-01 & -9.010480e-01 & -5.281260e-01 & -4.112000e-01\tabularnewline
7.787302e-01 & 1.000035e-01 & $\eta_{12}$ & 5.929010e-01 & 2.620350e-01 & -1.567570e-01 & -1.024260e-01\tabularnewline
\hline 
\end{tabular}
\end{sidewaystable*}

\begin{table}
\caption{\label{tab:GA results}Results of the genetic algorithm for the network
$y_{k}$ with lower energy}

\centering{}%
\begin{tabular}{c|c|c}
\hline 
Calculation & Variable & Value\tabularnewline
\hline 
\hline 
\multirow{10}{*}{Train} & $\alpha_{1}$ & 6.504147\tabularnewline
 & $\beta_{1}$ & -7.050345e-01\tabularnewline
 & $\gamma_{1}$ & 1.776158e-01\tabularnewline
 & $\chi_{1}$ & 1.752492e-01\tabularnewline
 & $E_{r}$ & 0.768\%\tabularnewline
 & $E$ & 5.969894e-02 enats\tabularnewline
 & $\left\langle T\right\rangle $ & 5.944988e-02 enats\tabularnewline
 & $\left\langle V\right\rangle $ & 2.490563e-04 enats\tabularnewline
 & $W$ & 2.772339 enats\tabularnewline
 & $\mathcal{C}$ & 8.982000e-05\tabularnewline
\hline 
\multirow{10}{*}{Test} & $\alpha_{1}$ & 7.058333\tabularnewline
 & $\beta_{1}$ & -5.941080e-01\tabularnewline
 & $\gamma_{1}$ & 1.418035e-01\tabularnewline
 & $\chi_{1}$ & 1.769226e-01\tabularnewline
 & $E_{r}$ & 0.782\%\tabularnewline
 & $E$ & 5.989879e-02 enats\tabularnewline
 & $\left\langle T\right\rangle $ & 5.964383e-02 enats\tabularnewline
 & $\left\langle V\right\rangle $ & 2.549622e-04 enats\tabularnewline
 & $W$ & 2.772333 enats\tabularnewline
 & $\mathcal{C}$ & 9.194974e-05\tabularnewline
\hline 
\end{tabular}
\end{table}

\begin{figure*}
\begin{centering}
\includegraphics[width=1\textwidth]{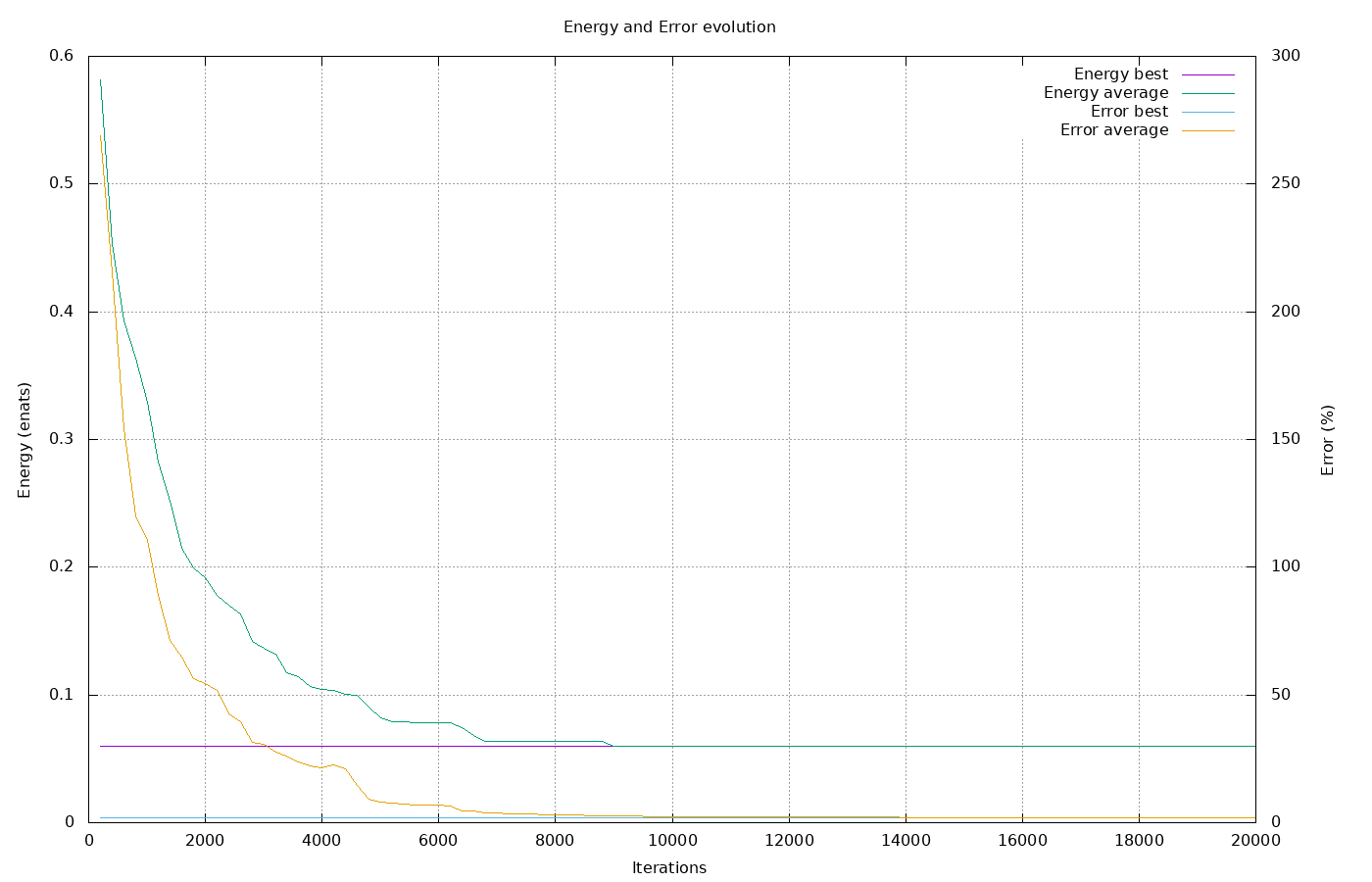}
\par\end{centering}
\caption{\label{fig:GA evolution}Evolution of the genetic algorithm that produced
the solution with lower energy for the dataset POLLEN. Average and
better (lower) energy and error decrease with the number of generations
of the GA.}
\end{figure*}

\begin{figure*}
\begin{centering}
\includegraphics[width=1\textwidth]{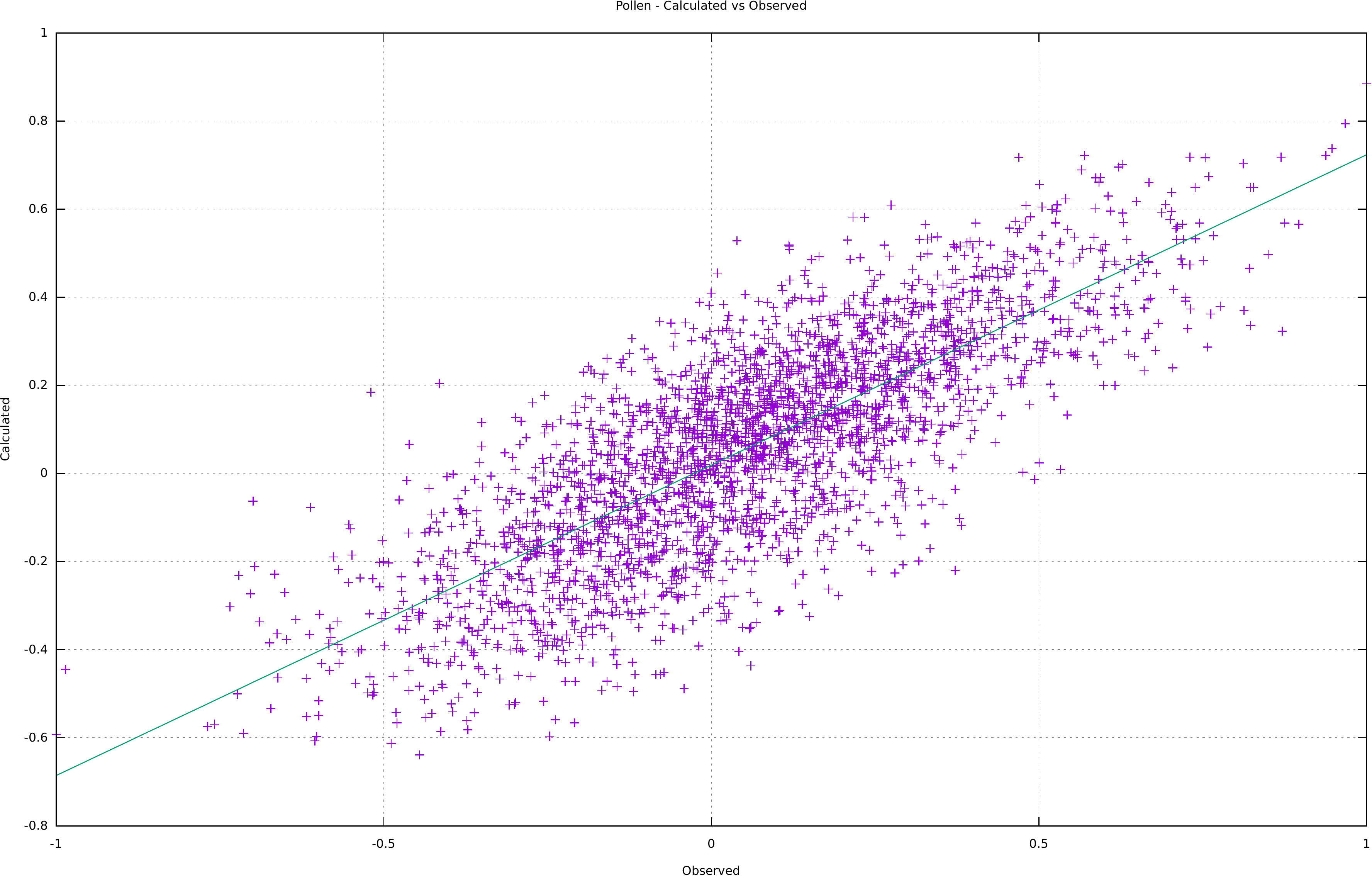}
\par\end{centering}
\caption{\label{fig:dispersion training}Dispersion plot for the POLLEN problem
of calculated data generated by the optimal neural network vs. observed
data (dataset) for the training partition.}
\end{figure*}

\begin{figure*}
\begin{centering}
\includegraphics[width=1\textwidth]{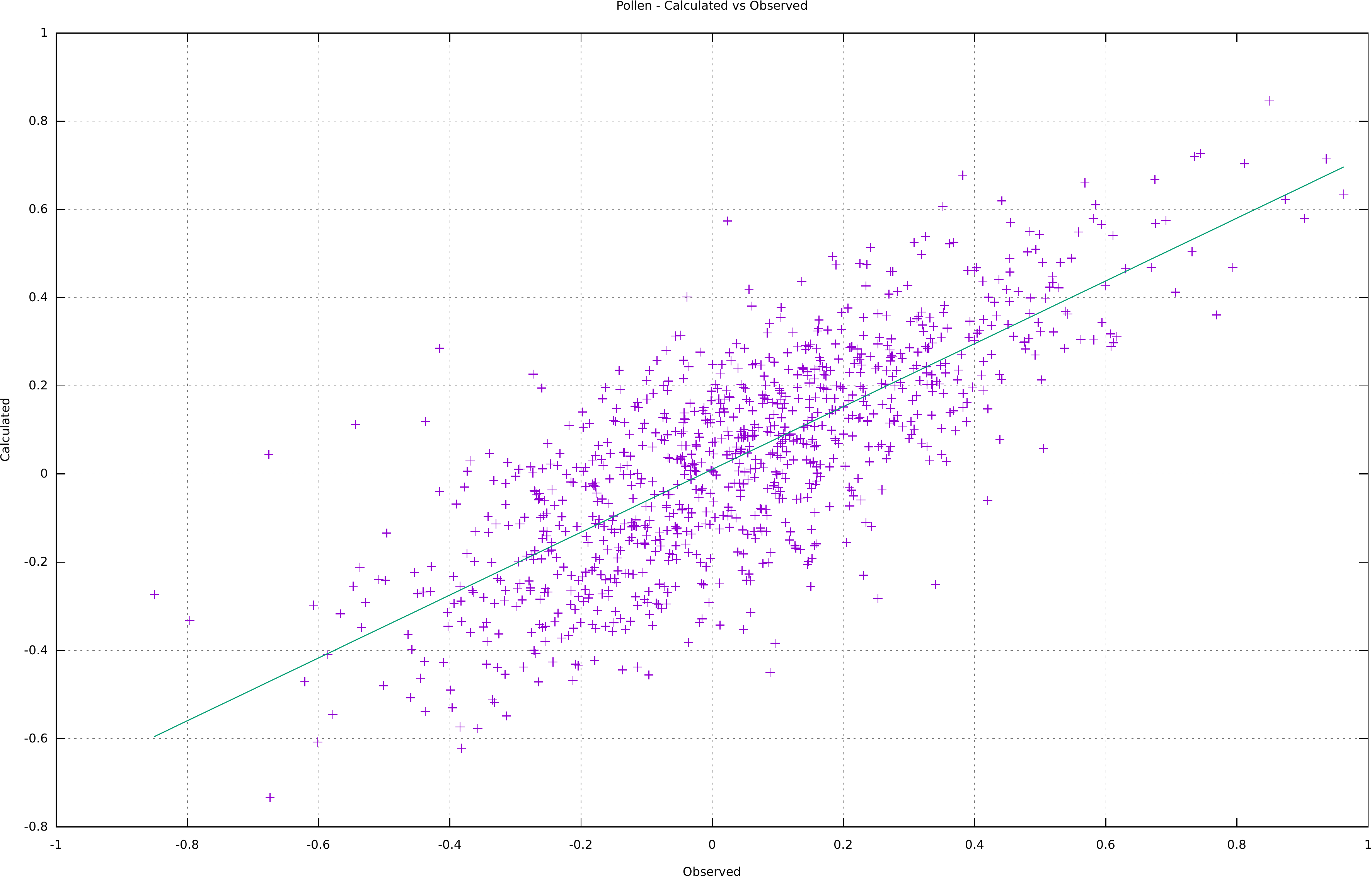}
\par\end{centering}
\caption{\label{fig:dispersion testing}Dispersion plot for the POLLEN problem
of calculated data generated by the optimal neural network vs. observed
data (dataset) for the testing partition.}
\end{figure*}

For each element of the population, in addition to the energy value,
has been calculated the square error percentage of the neural network
\citep{FranciscoYepesBarrera-2007-385,Prechelt:1994:PROBEN1}
\[
E_{r}=\frac{100}{s(t_{max}-t_{min})^{2}}\sum_{s}(y_{s}-t_{s})^{2}
\]
where $s$ is the number of records in the dataset and $t_{max}-t_{min}=2$
depends on the normalization interval used.

Some of the parameters in the Table \ref{tab:GA parameters} deserve
some observation:
\begin{itemize}
\item $\upsilon=1$ implies the so-called \emph{triangular niche sharing};
\item $R$ has a considerable influence on the results and was chosen for
each of the 10 concurrent processes of each calculation according
to the criterion $R_{i}=(i-1)/10,\,i=1,\ldots,10$, where $i$ is
the process number. This allows to avoid arbitrary choices since $R$
can be dependent on the nature of the problem;
\item $\xi$ was chosen in the interval $[0:4]$, which includes the value
given by a heuristic RBF rule which proposes for the standard deviation
of the associated normal distribution, $\sigma_{\xi}=\sqrt{\frac{1}{2\xi}}$,
the reference value $2\bar{d}_{\omega}$, where $\bar{d}_{\omega}$
is the average value between the centroids of the functions $\phi_{p}$
of the equation (\ref{eq:RBF function}). Considering an estimate
of $\bar{d}_{\omega}=1$ (half of the normalization interval) we get
$\xi=0.125$. The range $\xi\in[0:4]$ is equivalent to $\sigma_{\xi}\in[0.354:\infty]$.
The same criterion has also been used for the vector $\vec{\lambda}$.
\end{itemize}
The dataset was divided into two parts, a set of training (2886 records)
and a set of testing (962 records). Technically this subdivision is
not necessary, since the two data partitions are generated by the
same distribution and the characterization of the problem in the model
is given exclusively by the value of the constants $\vec{\rho}$,
$\vec{\theta}$, $\vec{\mu}$ and $\vec{\sigma}$, which are almost
the same for both partitions.

The variational parameters of the best solution are reported in Tables
\ref{tab:Net parameters}, \ref{tab:Weights} and \ref{tab:Psi parameters}.
The final results of the calculation, including error and energy,
are shown in Table \ref{tab:GA results}. Figure \ref{fig:GA evolution}
shows the evolution of error and energy (lower and average) of the
calculation that generated the lower energy solution, which shows
how the minimization of energy leads to a decrease of the error committed
by the net in the target prediction. The final error value for training
(0.768\%) and testing (0.782\%) partitions is particularly significant
given the low number of basis functions used in the definition of
$\Psi$ and $y_{k}$. Figures \ref{fig:dispersion training} and \ref{fig:dispersion testing}
show the dispersion graphs of the network output vs. observed values
for the target (dataset) generated by the optimal net.

\section{Conclusions}

In this work we have developed a model for optimizing artificial neural
networks based on an analogy with a physical quantum-mechanical system
using a potential based on the MinMI principle. This principle has
the sense to identify an approximation to the true relationship between
inputs and targets, stripping it of unnecessary superstructures. With
regard to the model that we have proposed, the starting point of this
paper is made from the observation of some analogies that lead to
treat the problem of the optimization of a neural network as a physical
system governed by an eigenvalue equation, taken without justification,
and to develop the consequences of such an approach.

The author took a lot of freedom in the initial setting of this work.
However, the quality of the results obtained in applying the model
to a series of problems (only one of these tests is reported in this
paper) prompted us to formalize the model and give it adequate mathematical
consistency. It is possible to deepen further the formal bases on
some of the topics treated and the mathematical expressions obtained,
many of which are postulated on the basis of simple dimensional analysis
of the equations. Extending the potential applicability of the formalism
and the results of quantum mechanics to the EANNs requires a separate
study that is outside the scope of this paper. Such a study is desirable
given the encouraging results obtained.

We underline some particular points of interest of this work:
\begin{enumerate}
\item The possibility of using, in the study of neural networks, the results
of a theory, quantum mechanics, with a high degree of mathematical
and conceptual maturity.
\item The potential possibility of applying the model to a wide variety
of problems, in particular those not covered in this paper, such as
datasets with discrete inputs/outputs and time series.
\item The fact that it is possible to realize the optimization of a neural
network for a problem by solving the system of linear equations (\ref{eq:Equations system on parameters}),
allowing in principle to obtain the solution with a deterministic
procedure in real time. If the system (\ref{eq:Equations system on parameters})
is calculable, then a training process intended in the sense of conventional
procedures such as backpropagation is not necessary. The potential
possibility of decrease the calculation times represents a point of
interest.
\item Results for the POLLEN problem is particularly significant,\footnote{At the time of this writing, only the calculations for the Pollen
dataset have been completed.} given the low final errors in the prediction of the dataset, obtained
with a very low number of basis functions in the definition of $\Psi$
and $y_{k}$.
\end{enumerate}
It is necessary to carry out a systematic test campaign to verify
the results obtained. These tests are currently underway and will
be the subject of a subsequent work. The preliminary results obtained
on a set of selected problems coming from the Statlib\footnote{\url{http: //lib.stat.cmu.edu/datasets/}}
and UCI\footnote{\url{https://archive.ics.uci.edu/ml/index.php}}
repositories confirm the validity of the model. Prediction errors
calculated with optimized EANNs are of the same order of magnitude
as those obtained with conventional techniques: genetic algorithms,
simulated annealing and backpropagation.

\settowidth{\nomlabelwidth}{$\mathcal{N}_{i}=\mathcal{N} (\mu_{i};\sigma_{i}^{2})$}
\printnomenclature{}

\nomenclature[N]{$N$}{dimension of input $\mathbf{x}$ (number of features $ x_ {i} $)}\nomenclature[y]{$y_{k}$}{network output for the $k$-th target}\nomenclature[t]{$\mathbf{t}$}{set of targets}\nomenclature[tk]{$t_{k}$}{$k$-th target}\nomenclature[x]{$\mathbf{x}$}{set of inputs (composed by features $x_{i}$}\nomenclature[xi]{$x_{i}$}{$i$-th input (feature)}\nomenclature[C]{$C$}{dimension of target $\mathbf{t}$}\nomenclature[Gamma]{$\Gamma$}{set of variational parameters}\nomenclature[Expected value]{$\left\langle \ldots \right\rangle$}{expected value}\nomenclature[V]{$V$}{potential}\nomenclature[E]{$E$}{energy}\nomenclature[Psi]{$\Psi$}{state function}\nomenclature[H]{$\hat{H}$}{hamiltonian operator}\nomenclature[F]{$\mathbf{F}$}{force}\nomenclature[epsilon]{$\epsilon$}{multiplicative constant of cinetic term of hamiltonian operator}\nomenclature[epsilon]{$I_{k}$}{mutual information fot the $k$-th net output}\nomenclature[chi]{$\chi_{k}$}{standard deviation between input a target for the $k$-th net output}\nomenclature[Sigma]{$\Sigma$}{covariance matrix}\nomenclature[thetak]{$\theta_{k}$}{standard deviation for $k$-th component of target $\mathbf{t}$}\nomenclature[rhok]{$\rho_{k}$}{centroid for $k$-th component of target $\mathbf{t}$}\nomenclature[sigmai]{$\sigma_{i}$}{standard deviation for $i$-th component of input $\mathbf{x}$}\nomenclature[mui]{$\mu_{i}$}{centroid for $i$-th component of input $\mathbf{x}$}\nomenclature[Ni]{$\mathcal{N}_{i}=\mathcal{N} (\mu_{i};\sigma_{i}^{2})$}{normal probability density with center $\mu_{i}$ and variance $\sigma_{i}^{2}$}\nomenclature[alphak]{$\alpha_{k}$}{constant in potential for $k$-th net output}\nomenclature[betak]{$\beta_{k}$}{constant in potential for $k$-th net output}\nomenclature[gammak]{$\gamma_{k}$}{constant in potential for $k$-th net output}\nomenclature[wkp]{$w_{kp}$}{weight between $p$-th basis function $\phi_{p}$ and $k$-th network output $y_{k}$}\nomenclature[phip]{$\phi_{p}$}{$p$-th basis function for network $y_{k}$}\nomenclature[xip]{$\xi_{p}$}{exponent in $p$-th basis function $\phi_{p}$}\nomenclature[omegapi]{$\omega_{pi}$}{centroid  in $p$-th basis function $\phi_{p}$ relative to $i$-th component of vector $\mathbf{x}$}\nomenclature[wk0]{$w_{k0}$}{bias for $k$-th network output $y_{k}$}\nomenclature[P]{$P$}{number of basis functions $\phi_{p}$}\nomenclature[Top]{$\hat{T}$}{cinetic energy operator}\nomenclature[Vop]{$\hat{V}$}{potential energy operator}\nomenclature[psid]{$\psi_{d}$}{$d$-th state for state function $\Psi$}\nomenclature[D]{$D$}{number of  states composing state function $\Psi$}\nomenclature[lambdad]{$\lambda_{d}$}{orbital exponent for state $\psi_{d}$}\nomenclature[etaid]{$\eta_{id}$}{centroid for state $\psi_{d}$ relative to $i$-th component of input $\mathbf{x}$}\nomenclature[Smn]{$S_{mn}$}{overlap integral for states $\phi_{m}$ and $\phi_{n}$}\nomenclature[Hmn]{$H_{mn}$}{hamiltonian integral for states $\phi_{m}$ and $\phi_{n}$}\nomenclature[cd]{$c_{d}$}{$d$-th coefficient of state function $\Psi$ relative to state $\psi_{d}$}\nomenclature[H]{$\mathbf{H}$}{hamiltonian matrix}\nomenclature[S]{$\mathbf{S}$}{overlap matrix}\nomenclature[w]{$\mathbf{w}$}{weight matrix}\nomenclature[p(a,b)]{$p(a\mid b)$}{conditional probability of $a$ given $b$}\nomenclature[p(a)]{$p(a)$}{marginal probability of $a$}\nomenclature[delta]{$\delta$}{Kronecker delta}\nomenclature[h(a)]{$h(a)$}{marginal differential entropy of $a$}\nomenclature[h(a,b)]{$h(a \mid b)$}{conditional differential entropy of $a$ given $b$}\nomenclature[W]{$W$}{work}\nomenclature[V0]{$V_{0}$}{potential at the optimum}\nomenclature[W0]{$W_{0}$}{work at the optimum}\nomenclature[Sc]{$\mathcal{S}$}{self-organization}\nomenclature[Ec]{$\mathcal{E}$}{emergency}\nomenclature[Cc]{$\mathcal{C}$}{complexity}\nomenclature[phivar]{$\varphi$}{function of diversity in niching mechanism}\nomenclature[dlm]{$d_{lm}$}{Hamming distance between chromosomes $l$ and $m$}\nomenclature[R]{$R$}{niche radius}\nomenclature[bi]{$b_{i}$}{binary code for $i$-th bit of chromosome}\nomenclature[gi]{$g_{i}$}{Gray code for $i$-th bit of chromosome}\nomenclature[XOR]{$\otimes$}{XOR operator}\nomenclature[Delta]{$\Delta$}{range of variability of dataset}\nomenclature[Er]{$E_{r}$}{square error percentage}\nomenclature[I]{$I(a,b)$}{Mutual information of $a$ and $b$}\nomenclature[I]{$\mathcal{I}$}{Information}

\bibliographystyle{plainnat}
\bibliography{/home/paco/bibliografia/bib_paco}

\begin{thebibliography}{82}
\providecommand{\natexlab}[1]{#1}
\providecommand{\url}[1]{\texttt{#1}}
\expandafter\ifx\csname urlstyle\endcsname\relax
  \providecommand{\doi}[1]{doi: #1}\else
  \providecommand{\doi}{doi: \begingroup \urlstyle{rm}\Url}\fi

\bibitem[del(2012)]{deloumeaux_information_2012}
Information theory: new research.
\newblock In Pierre Deloumeaux and Jose~D. Gorzalka, editors, \emph{Information
  theory: new research}, Mathematics research developments. Nova Science
  Publishers, New York, 2012.
\newblock ISBN 978-1-62100-325-0.

\bibitem[Angelow(2009)]{angelow_evolution_2009}
Andrew Angelow.
\newblock Evolution of {Schrödinger} {Uncertainty} {Relation} in {Quantum}
  {Mechanics}.
\newblock \emph{NeuroQuantology}, 7\penalty0 (2), February 2009.
\newblock ISSN 13035150.
\newblock \doi{10.14704/nq.2009.7.2.235}.

\bibitem[Baran et~al.(2001)Baran, Vallejos, Ramos, and
  Fernandez]{baran_multi-objective_2001}
B.~Baran, J.~Vallejos, R.~Ramos, and U.~Fernandez.
\newblock Multi-objective reactive power compensation.
\newblock In \emph{2001 {IEEE}/{PES} {Transmission} and {Distribution}
  {Conference} and {Exposition}. {Developing} {New} {Perspectives} ({Cat}.
  {No}.01CH37294)}, volume~1, pages 97--101, Atlanta, GA, USA, 2001. IEEE.
\newblock ISBN 978-0-7803-7285-6.
\newblock \doi{10.1109/TDC.2001.971215}.

\bibitem[Baran et~al.(2017)Baran, Harmancioglu, Cetinkaya, and
  Barbaros]{baran_extension_2017}
Turkay Baran, Nilgun~B. Harmancioglu, Cem~Polat Cetinkaya, and Filiz Barbaros.
\newblock An {Extension} to the {Revised} {Approach} in the {Assessment} of
  {Informational} {Entropy}.
\newblock \emph{Entropy}, 19\penalty0 (12):\penalty0 634, December 2017.
\newblock \doi{10.3390/e19120634}.

\bibitem[Barrera(2007)]{FranciscoYepesBarrera-2007-385}
Francisco~Yepes Barrera.
\newblock B{\'u}squeda de la estructura {\'o}ptima de redes neurales con
  algoritmos gen{\'e}ticos y simulated annealing. verificaci{\'o}n con el
  benchmark proben1.
\newblock \emph{Inteligencia Artificial, Revista Iberoamericana de IA},
  11\penalty0 (34):\penalty0 41--61, 2007.

\bibitem[Bastos et~al.(2015)Bastos, Bernardini, Bertolami, Dias, and
  Prata]{bastos_robertson-schrodinger_2015}
Catarina Bastos, Alex~E Bernardini, Orfeu Bertolami, Nuno~Costa Dias, and
  João~Nuno Prata.
\newblock Robertson-{Schrödinger} formulation of {Ozawa}'s uncertainty
  principle.
\newblock \emph{Journal of Physics: Conference Series}, 626:\penalty0 012050,
  July 2015.
\newblock ISSN 1742-6588, 1742-6596.
\newblock \doi{10.1088/1742-6596/626/1/012050}.

\bibitem[Beckner(1975)]{beckner_inequalities_1975}
William Beckner.
\newblock Inequalities in {Fourier} {Analysis}.
\newblock \emph{The Annals of Mathematics}, 102\penalty0 (1):\penalty0 159,
  July 1975.
\newblock ISSN 0003486X.
\newblock \doi{10.2307/1970980}.

\bibitem[Benoudjit and Verleysen(2003)]{benoudjit_kernel_2003}
Nabil Benoudjit and Michel Verleysen.
\newblock On the {Kernel} {Widths} in {Radial}-{Basis} {Function} {Networks}.
\newblock \emph{Neural Processing Letters}, 18:\penalty0 139--154, 2003.

\bibitem[Benoudjit et~al.(2002)Benoudjit, Archambeau, Lendasse, Lee, and
  Verleysen]{benoudjit_width_2002}
Nabil Benoudjit, C\'edric Archambeau, Amaury Lendasse, John Lee, and Michel
  Verleysen.
\newblock Width optimization of the {Gaussian} kernels in {Radial} {Basis}
  {Function} {Networks}.
\newblock In \emph{{ESANN} 2002 proceedings - {European} {Symposium} on
  {Artificial} {Neural} {Networks}}, pages 425--432, Belgium, 2002.
\newblock ISBN 2-930307-02-1.

\bibitem[Bialynicki-Birula and
  Mycielski(1975)]{bialynicki-birula_uncertainty_1975}
Iwo Bialynicki-Birula and Jerzy Mycielski.
\newblock Uncertainty relations for information entropy in wave mechanics.
\newblock \emph{Communications in Mathematical Physics}, 44\penalty0
  (2):\penalty0 129--132, June 1975.
\newblock ISSN 0010-3616, 1432-0916.
\newblock \doi{10.1007/BF01608825}.

\bibitem[Bishop(1995)]{Bishop:NNPR:1995}
Christopher~M. Bishop.
\newblock \emph{Neural Networks for Pattern Recognition}.
\newblock Oxford University Press, 1995.

\bibitem[Bourret(1958)]{bourret_note_1958}
Richard Bourret.
\newblock A note on an information theoretic form of the uncertainty principle.
\newblock \emph{Information and Control}, 1\penalty0 (4):\penalty0 398--401,
  December 1958.
\newblock ISSN 00199958.
\newblock \doi{10.1016/S0019-9958(58)90249-3}.

\bibitem[Braunstein and Caves(1990)]{braunstein_wringing_1990}
Samuel~L Braunstein and Carlton~M Caves.
\newblock Wringing out better {Bell} inequalities.
\newblock \emph{Annals of Physics}, 202\penalty0 (1):\penalty0 22--56, August
  1990.
\newblock ISSN 0003-4916.
\newblock \doi{10.1016/0003-4916(90)90339-P}.

\bibitem[Briggs(2008)]{briggs_derivation_2008}
J~S Briggs.
\newblock A derivation of the time-energy uncertainty relation.
\newblock \emph{Journal of Physics: Conference Series}, 99:\penalty0 012002,
  February 2008.
\newblock ISSN 1742-6596.
\newblock \doi{10.1088/1742-6596/99/1/012002}.

\bibitem[Cerf and Adami(1997)]{cerf_entropic_1997}
N.~J. Cerf and C.~Adami.
\newblock Entropic {Bell} {Inequalities}.
\newblock \emph{Physical Review A}, 55\penalty0 (5):\penalty0 3371--3374, May
  1997.
\newblock ISSN 1050-2947, 1094-1622.
\newblock \doi{10.1103/PhysRevA.55.3371}.

\bibitem[Chen et~al.(2008)Chen, Hu, Li, and Sun]{chen_adaptive_2008}
Badong Chen, Jinchun Hu, Hongbo Li, and Zengqi Sun.
\newblock Adaptive {FIR} {Filtering} under {Minimum} {Error}/{Input}
  {Information} {Criterion}.
\newblock \emph{IFAC Proceedings Volumes}, 41\penalty0 (2):\penalty0
  3539--3543, 2008.
\newblock ISSN 14746670.
\newblock \doi{10.3182/20080706-5-KR-1001.00598}.

\bibitem[Chen et~al.(2013)Chen, Zhu, Hu, and Principe]{chen_system_2013}
Badong Chen, Yu~Zhu, Jinchun Hu, and Jose~C. Principe.
\newblock System {Identification} {Based} on {Mutual} {Information} {Criteria}.
\newblock In \emph{System {Parameter} {Identification}}, pages 205--238.
  Elsevier, 2013.
\newblock ISBN 978-0-12-404574-3.
\newblock \doi{10.1016/B978-0-12-404574-3.00006-3}.

\bibitem[Chern(2017)]{chern_fluid_2017}
Albert Ren-Haur Chern.
\newblock \emph{Fluid {Dynamics} with {Incompressible} {Schr\"odinger} {Flow}}.
\newblock phd, California Institute of Technology, 2017.

\bibitem[Coles et~al.(2017)Coles, Berta, Tomamichel, and
  Wehner]{coles_entropic_2017}
Patrick~J. Coles, Mario Berta, Marco Tomamichel, and Stephanie Wehner.
\newblock Entropic {Uncertainty} {Relations} and their {Applications}.
\newblock \emph{Reviews of Modern Physics}, 89\penalty0 (1):\penalty0 015002,
  February 2017.
\newblock ISSN 0034-6861, 1539-0756.
\newblock \doi{10.1103/RevModPhys.89.015002}.

\bibitem[Contreras et~al.(2010)Contreras, Pellicer, Villena, and
  Ruiz]{contreras_quantum_2010}
Mauricio Contreras, Rely Pellicer, Marcelo Villena, and Aaron Ruiz.
\newblock A quantum model of option pricing: {When} {Black}-{Scholes} meets
  {Schr\"odinger} and its semi-classical limit.
\newblock \emph{Physica A: Statistical Mechanics and its Applications},
  389\penalty0 (23):\penalty0 5447--5459, December 2010.
\newblock ISSN 03784371.
\newblock \doi{10.1016/j.physa.2010.08.018}.

\bibitem[Dembo et~al.(1991)Dembo, Cover, and Thomas]{dembo_information_1991}
A.~Dembo, T.M. Cover, and J.A. Thomas.
\newblock Information theoretic inequalities.
\newblock \emph{IEEE Transactions on Information Theory}, 37\penalty0
  (6):\penalty0 1501--1518, November 1991.
\newblock ISSN 1557-9654.
\newblock \doi{10.1109/18.104312}.

\bibitem[DeWitt et~al.(1973)DeWitt, Everett, and
  Graham]{dewitt_many-worlds_1973}
Bryce~S. DeWitt, Hugh Everett, and Neill Graham.
\newblock \emph{The many-worlds interpretation of quantum mechanics: a
  fundamental exposition}.
\newblock Princeton series in physics. Princeton University Press, Princeton,
  N.J, 1973.
\newblock ISBN 978-0-691-08126-7 978-0-691-08131-1.

\bibitem[Falaye et~al.(2016)Falaye, Serrano, and Dong]{falaye_fishers_2016}
B.~J. Falaye, F.~A. Serrano, and Shi-Hai Dong.
\newblock Fisher's information for the position-dependent mass {Schr}\"odinger
  system.
\newblock \emph{Physics Letters A}, 380\penalty0 (1-2):\penalty0 267--271,
  January 2016.
\newblock ISSN 03759601.
\newblock \doi{10.1016/j.physleta.2015.09.029}.

\bibitem[Feldman and Crutchfield(1997)]{feldman_statistical_1997}
David~P. Feldman and James~P. Crutchfield.
\newblock Statistical {Measures} of {Complexity}: {Why}?
\newblock \emph{arXiv:cond-mat/9708186}, August 1997.
\newblock arXiv: cond-mat/9708186.

\bibitem[Fern{\'a}ndez et~al.(2013)Fern{\'a}ndez, Maldonado, and
  Gershenson]{fernandez_information_2013}
Nelson Fern{\'a}ndez, Carlos Maldonado, and Carlos Gershenson.
\newblock Information {Measures} of {Complexity}, {Emergence},
  {Self}-organization, {Homeostasis}, and {Autopoiesis}.
\newblock \emph{arXiv:1304.1842 [nlin, q-bio]}, April 2013.
\newblock arXiv: 1304.1842.

\bibitem[Finnegan and Song(2017)]{finnegan_maximum_2017}
Alex Finnegan and Jun~S. Song.
\newblock Maximum entropy methods for extracting the learned features of deep
  neural networks.
\newblock \emph{PLOS Computational Biology}, 13\penalty0 (10):\penalty0
  e1005836, October 2017.
\newblock ISSN 1553-7358.
\newblock \doi{10.1371/journal.pcbi.1005836}.

\bibitem[Fischer(2019)]{fischer_limiting_2019}
Andreas Fischer.
\newblock Limiting {Uncertainty} {Relations} in {Laser}-{Based} {Measurements}
  of {Position} and {Velocity} {Due} to {Quantum} {Shot} {Noise}.
\newblock \emph{Entropy}, 21\penalty0 (3):\penalty0 264, March 2019.
\newblock ISSN 1099-4300.
\newblock \doi{10.3390/e21030264}.

\bibitem[Fitzgerald et~al.(2011)Fitzgerald, Sincich, and
  Sharpee]{fitzgerald_minimal_2011}
Jeffrey~D. Fitzgerald, Lawrence~C. Sincich, and Tatyana~O. Sharpee.
\newblock Minimal {Models} of {Multidimensional} {Computations}.
\newblock \emph{PLOS Computational Biology}, 7\penalty0 (3):\penalty0 e1001111,
  March 2011.
\newblock ISSN 1553-7358.
\newblock \doi{10.1371/journal.pcbi.1001111}.

\bibitem[Flego et~al.(2012)Flego, Plastino, and Plastino]{flego_fisher_2012}
S~P Flego, A~Plastino, and A~R Plastino.
\newblock Fisher {Information} and {Quantum} {Mechanics}.
\newblock \emph{Applied Chemistry}, page~30, 2012.

\bibitem[Frank and Lieb(2012)]{frank_entropy_2012}
Rupert~L. Frank and Elliott~H. Lieb.
\newblock Entropy and the uncertainty principle.
\newblock \emph{Annales Henri Poincar\'e}, 13\penalty0 (8):\penalty0
  1711--1717, December 2012.
\newblock ISSN 1424-0637, 1424-0661.
\newblock \doi{10.1007/s00023-012-0175-y}.

\bibitem[Fr\"owis et~al.(2015)Fr\"owis, Schmied, and
  Gisin]{frowis_tighter_2015}
Florian Fr\"owis, Roman Schmied, and Nicolas Gisin.
\newblock Tighter quantum uncertainty relations following from a general
  probabilistic bound.
\newblock \emph{Physical Review A}, 92\penalty0 (1):\penalty0 012102, July
  2015.
\newblock ISSN 1050-2947, 1094-1622.
\newblock \doi{10.1103/PhysRevA.92.012102}.

\bibitem[Gao and Hu(2006)]{Gao:MTMBNGA:2006}
Lan Gao and Youwei Hu.
\newblock Multi-target matching based on niching genetic algorithm.
\newblock \emph{JCSNS International Journal of Computer Science and Network
  Security}, 6\penalty0 (7A), July 2006.

\bibitem[Gershenson and Fern{\'a}ndez(2012)]{gershenson_complexity_2012}
Carlos Gershenson and Nelson Fern{\'a}ndez.
\newblock Complexity and {Information}: {Measuring} {Emergence},
  {Self}-organization, and {Homeostasis} at {Multiple} {Scales}.
\newblock \emph{Complexity}, 18\penalty0 (2):\penalty0 29--44, November 2012.
\newblock ISSN 10762787.
\newblock \doi{10.1002/cplx.21424}.
\newblock arXiv: 1205.2026.

\bibitem[Gibilisco et~al.(2007)Gibilisco, Imparato, and
  Isola]{gibilisco_uncertainty_2007}
Paolo Gibilisco, Daniele Imparato, and Tommaso Isola.
\newblock Uncertainty principle and quantum {Fisher} information. {II}.
\newblock \emph{Journal of Mathematical Physics}, 48\penalty0 (7):\penalty0
  072109, July 2007.
\newblock ISSN 0022-2488, 1089-7658.
\newblock \doi{10.1063/1.2748210}.

\bibitem[Gibilisco et~al.(2008)Gibilisco, Imparato, and
  Isola]{gibilisco_robertson-type_2008}
Paolo Gibilisco, Daniele Imparato, and Tommaso Isola.
\newblock A {Robertson}-type uncertainty principle and quantum {Fisher}
  information.
\newblock \emph{Linear Algebra and its Applications}, 428\penalty0
  (7):\penalty0 1706--1724, April 2008.
\newblock ISSN 0024-3795.
\newblock \doi{10.1016/j.laa.2007.10.013}.

\bibitem[Globerson and Tishby(2004)]{globerson_minimum_2004}
Amir Globerson and Naftali Tishby.
\newblock The {Minimum} {Information} {Principle} for {Discriminative}
  {Learning}.
\newblock In \emph{Proceedings of the 20th {Conference} on {Uncertainty} in
  {Artificial} {Intelligence}}, {UAI} '04, pages 193--200, Arlington, Virginia,
  United States, 2004. AUAI Press.
\newblock ISBN 978-0-9749039-0-3.
\newblock event-place: Banff, Canada.

\bibitem[Globerson et~al.(2009)Globerson, Stark, Vaadia, and
  Tishby]{Globerson:2009:MIPNCA}
Amir Globerson, Eran Stark, Eilon Vaadia, and Naftali Tishby.
\newblock The minimum information principle and its application to neural code
  analysis.
\newblock \emph{Proceedings of the National Academy of Sciences of the United
  States of America, PNAS}, 106\penalty0 (9), march 2009.

\bibitem[Guia\c{s}u(1977)]{guiasu_information_1977}
Silviu Guia\c{s}u.
\newblock \emph{Information theory with applications}.
\newblock McGraw-Hill, New York, 1977.
\newblock ISBN 978-0-07-025109-0.

\bibitem[Hall(2004)]{hall_prior_2004}
Michael J.~W. Hall.
\newblock Prior information: {How} to circumvent the standard joint-measurement
  uncertainty relation.
\newblock \emph{Physical Review A}, 69\penalty0 (5):\penalty0 052113, May 2004.
\newblock ISSN 1050-2947, 1094-1622.
\newblock \doi{10.1103/PhysRevA.69.052113}.

\bibitem[Hilgevoord and Uffink(2016)]{hilgevoord_uncertainty_2016}
Jan Hilgevoord and Jos Uffink.
\newblock The {Uncertainty} {Principle}.
\newblock In Edward~N. Zalta, editor, \emph{The {Stanford} {Encyclopedia} of
  {Philosophy}}. Metaphysics Research Lab, Stanford University, winter 2016
  edition, 2016.

\bibitem[Hirschman(1957)]{hirschman_note_1957}
I.~I. Hirschman.
\newblock A {Note} on {Entropy}.
\newblock \emph{American Journal of Mathematics}, 79\penalty0 (1):\penalty0
  152--156, 1957.
\newblock ISSN 0002-9327.
\newblock \doi{10.2307/2372390}.

\bibitem[Hofer-Szab\'o(2019)]{hofer-szabo_quantum_2019}
G\'abor Hofer-Szab\'o.
\newblock Quantum mechanics as a noncommutative representation of classical
  conditional probabilities.
\newblock \emph{Journal of Mathematical Physics}, 60\penalty0 (6):\penalty0
  062106, June 2019.
\newblock ISSN 0022-2488, 1089-7658.
\newblock \doi{10.1063/1.5005578}.

\bibitem[Hradil and {\v R}eh{\'a}{\v c}ek(2004)]{hradil_uncertainty_2004}
Z.~Hradil and J.~{\v R}eh{\'a}{\v c}ek.
\newblock Uncertainty relations from {Fisher} information.
\newblock \emph{Journal of Modern Optics}, 51\penalty0 (6-7):\penalty0
  979--982, May 2004.
\newblock ISSN 0950-0340, 1362-3044.
\newblock \doi{10.1080/09500340410001663954}.

\bibitem[Husmeier(1997)]{husmeier_modelling_1997}
Dirk Husmeier.
\newblock \emph{Modelling {Conditional} {Probability} {Densities} with {Neural}
  {Networks}}.
\newblock PhD thesis, King's College London, University of London, 1997.

\bibitem[Klein(2019)]{klein_probabilistic_2019}
U.~Klein.
\newblock From probabilistic mechanics to quantum theory.
\newblock \emph{Quantum Studies: Mathematics and Foundations}, August 2019.
\newblock ISSN 2196-5609, 2196-5617.
\newblock \doi{10.1007/s40509-019-00201-w}.

\bibitem[Klein(2012)]{pahlavani_statistical_2012}
Ulf Klein.
\newblock A {Statistical} {Derivation} of {Non}-{Relativistic} {Quantum}
  {Theory}.
\newblock In Mohammad~Reza Pahlavani, editor, \emph{Measurements in {Quantum}
  {Mechanics}}. InTech, February 2012.
\newblock ISBN 978-953-51-0058-4.
\newblock \doi{10.5772/33075}.

\bibitem[Kurihara and Uyen~Quach(2015)]{kurihara_advantages_2015}
Yoshimasa Kurihara and Nhi~My Uyen~Quach.
\newblock Advantages of {Probability} {Amplitude} {Over} {Probability}
  {Density} in {Quantum} {Mechanics}.
\newblock \emph{Applied Physics Research}, 7\penalty0 (2):\penalty0 p66, March
  2015.
\newblock ISSN 1916-9647, 1916-9639.
\newblock \doi{10.5539/apr.v7n2p66}.

\bibitem[Leipnik(1959)]{leipnik_entropy_1959}
Roy Leipnik.
\newblock Entropy and the uncertainty principle.
\newblock \emph{Information and Control}, 2\penalty0 (1):\penalty0 64--79,
  April 1959.
\newblock ISSN 0019-9958.
\newblock \doi{10.1016/S0019-9958(59)90082-8}.

\bibitem[Levine(2014)]{levine_quantum_2014}
Ira~N. Levine.
\newblock \emph{Quantum chemistry}.
\newblock Pearson, Boston, seventh edition edition, 2014.
\newblock ISBN 978-0-321-80345-0.

\bibitem[Lopez-Ru{\'i}z et~al.(1995)Lopez-Ru{\'i}z, Mancini, and
  Calbet]{lopez-ruiz_statistical_1995}
Ricardo Lopez-Ru{\'i}z, Hector Mancini, and Xavier Calbet.
\newblock A statistical measure of complexity.
\newblock \emph{Physics Letters A}, 209\penalty0 (5):\penalty0 321--326,
  December 1995.
\newblock ISSN 0375-9601.
\newblock \doi{10.1016/0375-9601(95)00867-5}.

\bibitem[Madelung(1927)]{madelung_quantum_1927}
E.~Madelung.
\newblock Quantum {Theory} in {Hydrodynamical} {Form}.
\newblock \emph{Zeit. f. Phys.}, 40\penalty0 (322), 1927.

\bibitem[Majdisova and Skala(2017)]{majdisova_radial_2017}
Zuzana Majdisova and Vaclav Skala.
\newblock Radial {Basis} {Function} {Approximations}: {Comparison} and
  {Applications}.
\newblock \emph{Applied Mathematical Modelling}, 51:\penalty0 728--743,
  November 2017.
\newblock ISSN 0307904X.
\newblock \doi{10.1016/j.apm.2017.07.033}.

\bibitem[Majdisova and Skala(2018)]{majdisova_radial_2018}
Zuzana Majdisova and Vaclav Skala.
\newblock A {Radial} {Basis} {Function} {Approximation} for {Large} {Datasets}.
\newblock \emph{arXiv:1806.04243 [math]}, June 2018.

\bibitem[Messiah(1966)]{messiah_quantum_1966}
Albert Messiah.
\newblock \emph{Quantum {Mechanics}. {Two} volumes}.
\newblock North-Holland / NY: John Wiley \& Sons, fourth printing edition
  edition, 1966.

\bibitem[Movellan and McClelland(1993)]{movellan_learning_1993}
Javier~R. Movellan and James~L. McClelland.
\newblock Learning {Continuous} {Probability} {Distributions} with {Symmetric}
  {Diffusion} {Networks}.
\newblock \emph{Cognitive Science}, 17\penalty0 (4):\penalty0 463--496, October
  1993.
\newblock ISSN 03640213.
\newblock \doi{10.1207/s15516709cog1704-1}.

\bibitem[Norsen(2016)]{norsen_bohmian_2016}
Travis Norsen.
\newblock Bohmian {Conditional} {Wave} {Functions} (and the status of the
  quantum state).
\newblock \emph{Journal of Physics: Conference Series}, 701:\penalty0 012003,
  March 2016.
\newblock ISSN 1742-6588, 1742-6596.
\newblock \doi{10.1088/1742-6596/701/1/012003}.

\bibitem[Park and Abusalah(1997)]{park_maximum_1997}
Joseph~C. Park and Salahalddin~T. Abusalah.
\newblock Maximum {Entropy}: {A} {Special} {Case} of {Minimum} {Cross}-entropy
  {Applied} to {Nonlinear} {Estimation} by an {Artificial} {Neural} {Network}.
\newblock \emph{Complex Systems}, 11, 1997.

\bibitem[Parthasarathy(2009)]{parthasarathy_philosophy_2009}
K.~R. Parthasarathy.
\newblock On the philosophy of {Cram}{\textbackslash}'er-{Rao}-{Bhattacharya}
  {Inequalities} in {Quantum} {Statistics}.
\newblock \emph{arXiv:0907.2210 [cs, math, stat]}, July 2009.

\bibitem[Parwani(2005)]{parwani_why_2005}
Rajesh~R. Parwani.
\newblock Why is {Schr\"odinger}'s equation linear?
\newblock \emph{Brazilian Journal of Physics}, 35\penalty0 (2b):\penalty0
  494--496, June 2005.
\newblock ISSN 0103-9733.
\newblock \doi{10.1590/S0103-97332005000300021}.

\bibitem[Pires and Perdigao(2012)]{pires_minimum_2012}
Carlos A.~L. Pires and Rui A.~P. Perdigao.
\newblock Minimum {Mutual} {Information} and {Non}-{Gaussianity} {Through} the
  {Maximum} {Entropy} {Method}: {Theory} and {Properties}.
\newblock \emph{Entropy}, 14\penalty0 (6):\penalty0 1103--1126, June 2012.
\newblock ISSN 1099-4300.
\newblock \doi{10.3390/e14061103}.

\bibitem[Prechelt(1994)]{Prechelt:1994:PROBEN1}
Lutz Prechelt.
\newblock Proben1 - a set of neural network benchmark problems and benchmarking
  rules.
\newblock Technical Report 21/94, Fak\"ultat f\"ur Informatik, Universit\"at
  Karlsruhe, 76128 Karlsruhe, Germany, September 1994.

\bibitem[Reginatto(1999)]{reginatto_hydrodynamical_1999}
Marcel Reginatto.
\newblock Hydrodynamical formulation of quantum mechanics, {Kahler} structure,
  and {Fisher} information.
\newblock \emph{arXiv:quant-ph/9909065}, September 1999.

\bibitem[Robertson(1929)]{robertson_uncertainty_1929}
H.~P. Robertson.
\newblock The {Uncertainty} {Principle}.
\newblock \emph{Physical Review}, 34\penalty0 (1):\penalty0 163--164, July
  1929.
\newblock ISSN 0031-899X.
\newblock \doi{10.1103/PhysRev.34.163}.

\bibitem[Rodr\'iguez and Aguilar(2018)]{rodriguez_disturbance-disturbance_2018}
E.~Ben\'itez Rodr\'iguez and L.~M.~Ar\'evalo Aguilar.
\newblock Disturbance-{Disturbance} uncertainty relation: {The} statistical
  distinguishability of quantum states determines disturbance.
\newblock \emph{Scientific Reports}, 8\penalty0 (1):\penalty0 1--10, March
  2018.
\newblock ISSN 2045-2322.
\newblock \doi{10.1038/s41598-018-22336-3}.

\bibitem[Romero et~al.(2011)Romero, Gonzalez-Gaxiola, de~Chavez, and
  Bernal-Jaquez]{romero_black-scholes_2011}
Juan~M. Romero, O.~Gonzalez-Gaxiola, J.~Ruiz de~Chavez, and R.~Bernal-Jaquez.
\newblock The {Black}-{Scholes} {Equation} and {Certain} {Quantum}
  {Hamiltonians}.
\newblock \emph{arXiv:1002.1667 [hep-th, physics:math-ph, physics:quant-ph]},
  January 2011.

\bibitem[Rosenkrantz(1989)]{rosenkrantz_e._1989}
R.~D. Rosenkrantz, editor.
\newblock \emph{E. {T}. {Jaynes}: {Papers} on {Probability}, {Statistics} and
  {Statistical} {Physics}}.
\newblock Springer, Dordrecht, 1989 edition edition, April 1989.
\newblock ISBN 978-0-7923-0213-1.

\bibitem[Santamar{\'i}a-Bonfil et~al.(2016)Santamar{\'i}a-Bonfil,
  Fern{\'a}ndez, and Gershenson]{santamaria-bonfil_measuring_2016}
Guillermo Santamar{\'i}a-Bonfil, Nelson Fern{\'a}ndez, and Carlos Gershenson.
\newblock Measuring the {Complexity} of {Continuous} {Distributions}.
\newblock \emph{Entropy}, 18\penalty0 (3):\penalty0 72, February 2016.
\newblock ISSN 1099-4300.
\newblock \doi{10.3390/e18030072}.
\newblock arXiv: 1511.00529.

\bibitem[Stam(1959)]{stam_inequalities_1959}
A.~J. Stam.
\newblock Some inequalities satisfied by the quantities of information of
  {Fisher} and {Shannon}.
\newblock \emph{Information and Control}, 2\penalty0 (2):\penalty0 101--112,
  June 1959.
\newblock ISSN 0019-9958.
\newblock \doi{10.1016/S0019-9958(59)90348-1}.

\bibitem[Stepanov and Levitov(2001)]{stepanov_laplacian_2001}
M.~G. Stepanov and L.~S. Levitov.
\newblock Laplacian growth with separately controlled noise and anisotropy.
\newblock \emph{Physical Review E}, 63\penalty0 (6):\penalty0 061102, May 2001.
\newblock ISSN 1063-651X, 1095-3787.
\newblock \doi{10.1103/PhysRevE.63.061102}.

\bibitem[Steuer et~al.(2005)Steuer, Daub, Selbig, and
  Kurths]{steuer_measuring_2005}
Ralf Steuer, Carsten~O. Daub, Joachim Selbig, and J\"urgen Kurths.
\newblock Measuring {Distances} {Between} {Variables} by {Mutual}
  {Information}.
\newblock In Daniel Baier and Klaus-Dieter Wernecke, editors, \emph{Innovations
  in {Classification}, {Data} {Science}, and {Information} {Systems}}, Studies
  in {Classification}, {Data} {Analysis}, and {Knowledge} {Organization}, pages
  81--90, Berlin, Heidelberg, 2005. Springer.
\newblock ISBN 978-3-540-26981-6.
\newblock \doi{10.1007/3-540-26981-9_11}.

\bibitem[Theodoridis(2015)]{theodoridis_machine_2015}
Sergios Theodoridis.
\newblock \emph{Machine learning: a {Bayesian} and optimization perspective}.
\newblock Elsevier, AP, Amsterdam Boston Heidelberg London New York Oxford
  Paris San Diego San Francisco Singapore Sydney Tokyo, 2015.
\newblock ISBN 978-0-12-801522-3.

\bibitem[Tsekov(2012)]{tsekov_bohmian_2012}
R.~Tsekov.
\newblock Bohmian mechanics versus {Madelung} quantum hydrodynamics.
\newblock \emph{arXiv:0904.0723 [cond-mat, physics:quant-ph]}, 2012.
\newblock \doi{10.13140/RG.2.1.3663.8245}.

\bibitem[Tsekov et~al.(2019)Tsekov, Heifetz, and
  Cohen]{tsekov_hydrodynamic_2019}
Roumen Tsekov, Eyal Heifetz, and Eliahu Cohen.
\newblock A {Hydrodynamic} {Interpretation} of {Quantum} {Mechanics} via
  {Turbulence}.
\newblock \emph{arXiv:1804.00395 [physics, physics:quant-ph]}, May 2019.
\newblock \doi{10.7546/CRABS.2019.04.03}.

\bibitem[Vadasz(2016)]{vadasz_rendering_2016}
Peter Vadasz.
\newblock Rendering the {Navier}-{Stokes} {Equations} for a {Compressible}
  {Fluid} into the {Schr\"odinger} {Equation} for {Quantum} {Mechanics}.
\newblock \emph{Fluids}, 1\penalty0 (2):\penalty0 18, June 2016.
\newblock \doi{10.3390/fluids1020018}.

\bibitem[Villaverde et~al.(2014)Villaverde, Ross, Mor\'an, and
  Banga]{villaverde_mider_2014}
Alejandro~F. Villaverde, John Ross, Federico Mor\'an, and Julio~R. Banga.
\newblock {MIDER}: {Network} {Inference} with {Mutual} {Information} {Distance}
  and {Entropy} {Reduction}.
\newblock \emph{PLoS ONE}, 9\penalty0 (5):\penalty0 e96732, May 2014.
\newblock ISSN 1932-6203.
\newblock \doi{10.1371/journal.pone.0096732}.

\bibitem[Wajih et~al.(2017)Wajih, Sifaoui, and
  Abdelkrim]{wajih_logarithmic_2017}
Mohamed Wajih, Amel Sifaoui, and Afef Abdelkrim.
\newblock Logarithmic {Spiral}-based {Construction} of {RBF} {Classifiers}.
\newblock \emph{International Journal of Advanced Computer Science and
  Applications}, 8\penalty0 (2), 2017.
\newblock ISSN 21565570, 2158107X.
\newblock \doi{10.14569/IJACSA.2017.080235}.

\bibitem[Wu et~al.(2012)Wu, Wang, Zhang, and Du]{wu_using_2012}
Yue Wu, Hui Wang, Biaobiao Zhang, and K.-L. Du.
\newblock Using {Radial} {Basis} {Function} {Networks} for {Function}
  {Approximation} and {Classification}.
\newblock \emph{ISRN Applied Mathematics}, 2012:\penalty0 1--34, 2012.
\newblock ISSN 2090-5572.
\newblock \doi{10.5402/2012/324194}.

\bibitem[Xi et~al.(2018)Xi, Tang, and Luo]{xi_feature-level_2018}
Xugang Xi, Minyan Tang, and Zhizeng Luo.
\newblock Feature-{Level} {Fusion} of {Surface} {Electromyography} for
  {Activity} {Monitoring}.
\newblock \emph{Sensors}, 18\penalty0 (2):\penalty0 614, February 2018.
\newblock ISSN 1424-8220.
\newblock \doi{10.3390/s18020614}.

\bibitem[Xiaodong(2014)]{xiaodong_evaluation_2014}
Zhang Xiaodong.
\newblock Evaluation model and simulation of basketball teaching quality based
  on maximum entropy neural network.
\newblock page~5, 2014.

\bibitem[Xu(1999)]{Xu:1999}
Dongxin Xu.
\newblock \emph{Energy, entropy and information potential for neural
  computation}.
\newblock PhD thesis, University of Florida, 1999.

\bibitem[Yuan and Parwani(2009)]{yuan_properties_2009}
Liew~Ding Yuan and Rajesh~R Parwani.
\newblock Properties of some nonlinear {Schr\"odinger} equations motivated
  through information theory.
\newblock \emph{Journal of Physics: Conference Series}, 174:\penalty0 012043,
  June 2009.
\newblock ISSN 1742-6596.
\newblock \doi{10.1088/1742-6596/174/1/012043}.

\bibitem[Zhang et~al.(2017)Zhang, Ozay, Sun, and
  Okatani]{zhang_information_2017}
Yan Zhang, Mete Ozay, Zhun Sun, and Takayuki Okatani.
\newblock Information {Potential} {Auto}-{Encoders}.
\newblock \emph{arXiv:1706.04635 [cs, math, stat]}, June 2017.
\newblock arXiv: 1706.04635.

\end{thebibliography}

\end{document}